\documentclass[10pt, a4paper]{article}
\usepackage{lrec2022} 
\usepackage{multibib}
\newcites{languageresource}{Language Resources}
\usepackage{graphicx}
\usepackage{tabularx}
\usepackage{soul}

\usepackage{multirow}
\usepackage{caption}
\usepackage{subcaption}
\usepackage{todonotes}
\usepackage{listings}
\usepackage{tabularx}
\usepackage{float}
\usepackage{tabularx}
\usepackage{booktabs}
\usepackage{svg}
\usepackage{enumitem}

\usepackage{titlesec}
\titleformat{\section}{\normalfont\large\bfseries\center}{\thesection.}{1em}{}
\titleformat{\subsection}{\normalfont\SmallTitleFont\bfseries\raggedright}{\thesubsection.}{1em}{}
\titleformat{\subsubsection}{\normalfont\normalsize\bfseries\raggedright}{\thesubsubsection.}{1em}{}
\renewcommand\thesection{\arabic{section}}
\renewcommand\thesubsection{\thesection.\arabic{subsection}}
\renewcommand\thesubsubsection{\thesubsection.\arabic{subsubsection}}

\usepackage{epstopdf}
\usepackage[utf8]{inputenc}

\usepackage{hyperref}
\usepackage{xstring}

\usepackage{color}

\title{EmoWOZ: A Large-Scale Corpus and Labelling Scheme for Emotion Recognition in Task-Oriented Dialogue Systems}

\name{Shutong Feng, Nurul Lubis, Christian Geishauser, Hsien-chin Lin, \\
        {\bf \large Michael Heck, Carel van Niekerk and Milica Ga\v{s}i\'{c}}}

\address{Heinrich Heine University Düsseldorf\ \\
         Universitätsstraße 1, 40225 Düsseldorf, Germany \\
         \{fengs, lubis, geishaus, linh, heckmi, niekerk, gasic\}@hhu.de\\}

\abstract{The ability to recognise emotions lends a conversational artificial intelligence a human touch. While emotions in chit-chat dialogues have received substantial attention, emotions in task-oriented dialogues remain largely unaddressed. This is despite emotions and dialogue success having equally important roles in a natural system. Existing emotion-annotated task-oriented corpora are limited in size, label richness, and public availability, creating a bottleneck for downstream tasks. To lay a foundation for studies on emotions in task-oriented dialogues, we introduce EmoWOZ, a large-scale manually emotion-annotated corpus of task-oriented dialogues. EmoWOZ is based on MultiWOZ, a multi-domain task-oriented dialogue dataset. It contains more than 11K dialogues with more than 83K emotion annotations of user utterances. In addition to Wizard-of-Oz dialogues from MultiWOZ, we collect human-machine dialogues within the same set of domains to sufficiently cover the space of various emotions that can happen during the lifetime of a data-driven dialogue system. To the best of our knowledge, this is the first large-scale open-source corpus of its kind. We propose a novel emotion labelling scheme, which is tailored to task-oriented dialogues. We report a set of experimental results to show the usability of this corpus for emotion recognition and state tracking in task-oriented dialogues. \\ \newline  }

\begin{document}

\maketitleabstract

\section{Introduction}
Incorporating human intelligence into conversational artificial intelligence (AI) has been a challenging and long-term goal~\cite{Picard97}. Emotional intelligence, defined as the ability to regulate, perceive, assimilate, and express emotions, is a key component of general intelligence~\cite{MAYER1999267}. Such emotion awareness can help the conversational AI generate more emotionally and semantically appropriate responses~\cite{10.5555/3504035.3504125}.\par

Dialogue systems generally fall into two classes. Task-oriented systems converse with users to help complete tasks determined by user goals. Chit-chat systems are set up to mimic the unstructured conversations or ‘chats’ characteristic of human-human interaction~\cite{10.5555/1214993}. Chat-oriented systems are typically modelled in a supervised fashion with large available corpora~\cite{vinyal-etal-2015}. In contrast, task-oriented systems track the user goal throughout the dialogue and a policy is typically trained via some form of reinforcement learning (RL) to conduct dialogue towards successful goal completion~\cite{young2002talking}. Moreover, the scope of the dialogue can also be extended during this process, e.g.\ by adding new domains to the dialogue system~\cite{madotto-etal-2021-continual}. Consequently, the distribution of data from which a task-oriented system learns can change.

Emotions appear in both chit-chat and task-oriented dialogues. However, the cause of emotion may differ as well as their role. Chit-chat dialogues are a means to express emotion. Speakers may discuss emotional experiences~\cite{li-etal-2017-dailydialog}, or topics that induce emotions such as news broadcasts~\cite{8273582}. In task-oriented dialogues, the user is primarily interested in achieving their goal. While an emotional situation may be a reason to interact with the system, e.g.\ the user just missed a flight and needs to rebook one, the emotion the user exhibits is more often a reaction to potential goal completion or failure. Since the emotion is centred around the user goal, it is more contextual and subtle. Therefore, besides inferring emotional states from dialogue utterances, an agent also needs to reason about emotion-generating situations \cite{Poria2021}. \par 

Substantial research efforts in emotion recognition in conversations (ERC) have been invested in chit-chat dialogues~\cite{Majumder_Poria_Hazarika_Mihalcea_Gelbukh_Cambria_2019,ghosal-etal-2020-cosmic}. There are several public ERC corpora containing chit-chat dialogues~\cite{li-etal-2017-dailydialog,poria-etal-2019-meld,zahiri:18a} and conversational data from social media~\cite{zhou-wang-2018-mojitalk}. These corpora can tremendously accelerate the building of emotional chatbots using data-driven approaches~\cite{10.5555/3504035.3504125}. In task-oriented dialogues, recognising emotions is equally important but remains largely unaddressed. Using RL to optimise a dialogue policy necessitates a feedback signal. While it is accepted that the feedback signal needs to correlate with user satisfaction~\cite{Ultes2017DomainIndependentUS}, this feedback signal is often based on hand-coded rules. Could an emotional model instead be directly used to provide such a feedback signal? Could it also be used to support emotion-aware natural language generation~\cite{mairesse-walker-2007-personage}, or even improve dialogue state tracking through multi-task learning~\cite{heck-etal-2020-task}? Existing corpora are small in size, and labels are limited to sentiment polarity, creating a bottleneck, so these questions remain largely unexplored.\par

%
In this work, we present \textbf{EmoWOZ}, a large-scale manually labelled corpus for emotion in task-oriented dialogues. EmoWOZ is derived from MultiWOZ~\cite{budzianowski-etal-2018-multiwoz}, one of the largest multi-domain corpora and the benchmark dataset for various dialogue modelling tasks, from dialogue state tracking \cite{heck-etal-2020-trippy,lin-etal-2021-knowledge} to policy optimisation \cite{he2022galaxy}. We also collected and annotated human-machine dialogues as a complement. Our contributions are as follows: \par
\begin{itemize}[leftmargin=*]
    \item We construct a corpus containing task-oriented dialogues with emotion labels,
    comprising more than 11K dialogues and 83K annotated user utterances. To the best of our knowledge, this is the first large-scale open-source corpus \& code\footnote{https://doi.org/10.5281/zenodo.5865437} for emotion recognition in task-oriented dialogues.
    \item We propose a novel labelling scheme, containing 7 emotion classes, adapted from the Ortony, Clore and Collins (OCC) model~\cite{ortony_clore_collins_1988}, specifically tailored to capture an array of emotions in relation to user goals in task-oriented dialogue. 
    \item We report a series of emotion recognition baseline results to show the usability of this corpus. We also empirically show that the emotion labels can be used to improve the performance of other task-oriented dialogue system modules, in this case, a dialogue state tracker (DST).
\end{itemize}

\section{Related Work}

\begin{table*}[t]
\centering
\footnotesize
\setlength \tabcolsep{4pt}
\begin{tabular}{l|rrr|rrrr}
\toprule[1pt]
\multicolumn{1}{c|}{\textbf{Metric}} & \textbf{DailyDialog} & \textbf{MELD} & \textbf{EmoryNLP} & \textbf{DSTC1} & \textbf{SentiVA} & \textbf{TML} & \textbf{EmoWOZ(Ours)}\\
\hline 
Dialogue type & \multicolumn{3}{c|}{Chit-chat} & \multicolumn{4}{c}{Task-oriented} \\
\# Dialogues & \textbf{13,118} & 1,433 & 897 & 50 & 1,282 & 3,496 & 11,434 \\
Total \# turns & 102,979 & 13,708 & 12,606 & 517 & 35,267 & 68,216 & \textbf{167,234}\\
\# Unique tokens & 26,364 & 8052 & 8441 & 199 & - & - & \textbf{28,417} \\
Avg. turns / dialogue & 7.9 & 9.6 & 14.1 & 10.3 & \textbf{27.5} & 19.5 & 14.63\\
Avg. tokens / turn & \textbf{14.6} & 10.4 & 14.3 & 2.3 & - & - & 12.78 \\
Label type & Emo & \textbf{Sent, Emo} & \textbf{Sent, Emo} & Sent & Sent & Sent & \textbf{Sent, Emo} \\
\# Classes & 7 & \textbf{3 and 7} & \textbf{3 and 7} & 3 & 3 & 5 & \textbf{3 and 7} \\
\# Annotations & \textbf{102,879} & 13,708 & 12,606 & 517 & 35,267 & 68,216 & 83,617 \\
Neutral Samples (\%) & 83.1\% & 47.0\% & \textbf{30.0\%} & - & 88.6\% & 45.7\% & 70.1\% \\
\# Annotators / turn & 3 & 3 & \textbf{4} & - & 3 & 2 & 3 \\
Expert Annotator? & Yes & No & No & - & No & No & No \\
Agreement & 0.789 & 0.43 & 0.14 & - & \textbf{0.8} & 0.79 & 0.602 \\
Open-sourced? & \textbf{Yes} & \textbf{Yes} & \textbf{Yes} & \textbf{Yes} & No & No & \textbf{Yes} \\
\bottomrule[1pt]
\end{tabular}
\caption{Comparison of our corpus to similar corpora. Values in bold indicate the best value for each metric. For label type, ``Emo'' stands for emotion categories and ``Sent'' stands for sentiment polarities. For corpora providing both emotion and sentiment labels, agreement metrics are measured for emotion labels. DSTC1, SentiVA, and TML refer to works by \protect\newcite{shi-yu-2018-sentiment}, \protect\newcite{10.1371/journal.pone.0235367}, and \protect\newcite{Wang_Wang_Sun_Li_Liu_Si_Zhang_Zhou_2020}, respectively.}
\label{tab:dataset-comparison}
\vspace*{-3mm}
\end{table*}

\subsection{Emotion Models}
Within the area of affective computing, emotion models are commonly grouped into two types: dimensional models and categorical models.

\textbf{Dimensional models} describe emotions as a combination of values across a set of dimensions. The longest established dimensions are valence and arousal, as proposed by \newcite{russell1980circumplex} in the circumplex model of emotion. Valence measures the positivity, while arousal measures the activation. Happiness, for example, is an emotion with positive valence and high activation. Additional dimensions, namely dominance and expectancy \cite{fontaine2007world}, have also been proposed to further describe and distinguish complex emotions.\par

\textbf{Categorical models} group emotions into distinct categories. The ``Big six'' theory is one of the most well-known theories on universal emotions. Based on studies of facial expressions, \newcite{Ekman92anargument} proposed six basic human emotions which are influenced neither by culture nor other social influences: happiness, anger, sadness, disgust, fear, surprise.  \newcite{2001Eisp} conceptualised over a hundred emotions into a tree-structured list and identified six primary emotions from it.

\newcite{ortony_clore_collins_1988} proposed the Ortony, Clore and Collins (OCC) emotion model, which is explicitly developed for implementation in computers. In the OCC model, 22 emotion types are described as a valenced reaction to one of three cognitive elicitors: consequences of events, actions of agents, or aspects of objects. For example, \textit{dissatisfied} is specified as disapproving of someone else's blameworthy action. These cognitive aspects are in line with the cognitive process of a computational agent, making the OCC model suitable for building emotional artificial agents. However, the use of this model for dialogue agents is not yet wide-spread. In a similar spirit, \newcite{grossjjemotionregulation} formulated the process of emotion regulation as the attention, appraisal, and response originated from various situations. \par

Although there are corpora with real-valued annotation of multiple emotion dimensions~\cite{preotiuc-pietro-etal-2016-modelling,buechel-hahn-2017-emobank}, researchers often focus on the valence dimension and annotate with discrete classes~\cite{socher-etal-2013-recursive}, often called sentiment polarity. Emotion datasets also consider emotions from various categorical models in the annotation scheme~\cite{li-etal-2017-dailydialog,poria-etal-2019-meld}, but some datasets have domain-specific labels. For instance, \newcite{zhou-wang-2018-mojitalk} leverage common emojis in social media posts. The Topical-Chat dataset~\cite{Gopalakrishnan2019} introduces \textit{curious to dive deeper} in addition to other basic emotions.\par

In this work, we propose a novel set of 7 emotions and motivate it using OCC model as the basis. We aim for this scheme to capture the cognitive context of emotions while retaining the simplicity of labels that facilitates large-scale crowd-sourcing of emotion annotations.\par

\subsection{Emotion Dialogue Datasets}
Early works on ERC focus on speech signals \cite{911197,Riccardi2005GroundingEI,Carrin2008InfluenceOC}. More recently, there are increasing number of text-based ERC datasets focusing on chit-chat dialogue. Chit-chat dialogue lends itself well to affective computing research due to its open-domain set-up, where conversation topics are diverse and not restricted to a particular task. One of the largest such corpora is DailyDialog~\cite{li-etal-2017-dailydialog}, which contains conversations between English learners on various topics ranging from relationships to money. Other similar datasets include EmoryNLP~\cite{zahiri:18a} and MELD~\cite{poria-etal-2019-meld}. They contain multi-party dialogues from the TV show \textit{Friends}. TV recordings in talk show format have also been utilised to collect emotion-rich and topic-specific dialogues \cite{lubis2015construction}. Unfortunately, existing data suitable for task-oriented corpora, such as customer service chat logs, are typically not within the public domain.\par

There also exist a few corpora concerning the affective aspect of task-oriented dialogues. \newcite{Wang_Wang_Sun_Li_Liu_Si_Zhang_Zhou_2020} proposed a large-scale sentiment classification corpus containing customer service dialogues in Chinese. However, this dataset is not publicly available. \newcite{10.1371/journal.pone.0235367} annotated dialogues from bAbI~\cite{BordesBW17} with sentiment for policy optimisation. Since dialogues are machine-generated, it is unclear how well these emotions match real human emotions and whether sentiment on its own sufficiently captures emotional nuances in task-oriented dialogue. In a similar spirit, \newcite{shi-yu-2018-sentiment} annotated the DSTC1 dataset with user sentiment. Unfortunately, containing only 50 dialogues, the dataset is very limited in terms of coverage and application in machine learning. To summarise, existing corpora are either limited in size or not publicly available, limiting further works on emotions in task-oriented dialogue systems. Furthermore, their annotation schemes focus on sentiment polarities, overlooking the effect of goals on users' emotional states. 

\section{Dataset Construction}
\subsection{Task-oriented Dialogues}
\textbf{MultiWOZ:} Our dataset covers the entirety of MultiWOZ, which was constructed using the Wizard-of-Oz framework~\cite{10.1145/357417.357420}. It contains over 10k dialogues. Each dialogue was completed by two workers, each acting as the user or the operator, to achieve specified goals such as information retrieval or making reservations. There are 7 domains in total. A single dialogue or even a single turn can span multiple domains.\par 

\textbf{Complementary Dialogues:}
We envisage emotions as learning signal for dialogue system optimisation. Since emotions in task-oriented dialogue systems can be a direct effect of the user perception of the ability of the system to fulfill their goal, the policy performance can largely influence emotion distribution. During the life span of a data-driven task-oriented dialogue system, the distributions of dialogues and emotions may change as the policy learns and improves over time. An immediate impact of such a distributional shift is the increase in the number of negative emotions due to failed dialogues during the early stages of learning. Therefore, in addition to the human wizard policy in MultiWOZ, it is important that EmoWOZ covers a variety of dialogues which represent the emotions throughout such a dialogue system life span.

We complement MultiWOZ with human-machine \textbf{dial}ogues from a \textbf{ma}chine-\textbf{ge}nerated policy (\textbf{DialMAGE}). To elicit more genuine reactions, we let subjects directly interact with a machine-generated policy instead of human wizards trying to make machine-like mistakes. We launched a dialogue interactive task on Amazon Mechanical Turk, where workers are asked to retrieve information by interacting with the learning policy. We start with a policy trained in a supervised fashion on MultiWOZ 
that achieved a task success rate of 55\% when evaluated with the ConvLab-2 \cite{convlab2} rule-based user simulator. Throughout the task, the policy learned and improved as user feedback on task success is used for further training using RL. The policy reached a final human-rated success rate of 73\%. Similar to \newcite{li2020}, the policy uses a recurrent neural network (RNN) based model to produce multiple actions in a single turn, followed by the ConvLab-2 template-based NLG module for response generation. \par

\subsection{Emotion Annotation Scheme}

\begin{table*}
\centering
\scriptsize
\setlength\tabcolsep{3pt}
\renewcommand{\arraystretch}{1.1}
\begin{tabular}{ccclll}
\toprule[1pt]
\textbf{Elicitor} & \textbf{Valence} & \textbf{Conduct} & \multicolumn{1}{c}{\textbf{OCC Emotion}} & \multicolumn{1}{c}{\textbf{Our Emotion}} & \multicolumn{1}{c}{\textbf{Implication of User}} \\ \hline
 &  & Polite &  & {\color[HTML]{000000} \textcolor{blue}{Satisfied}, liking, appreciative} & Satisfied with the operator because the goal is fulfilled. \\ \cline{3-3} \cline{5-6} 
 & \multirow{-2}{*}{Positive} & Impolite & \multirow{-2}{*}{Admiration, gratitude, love} & \multicolumn{2}{l}{{\color[HTML]{9B9B9B} Not applicable to the dataset}} \\ \cline{2-6} 
 &  & Polite &  & \textcolor{blue}{Dissatisfied}, disliking & Dissatisfied with the operator's suggestion or mistake. \\ \cline{3-3} \cline{5-6} 
\multirow{-4}{*}{Operator} & \multirow{-2}{*}{Negative} & Impolite & \multirow{-2}{*}{Reproach, anger, hate} & \textcolor{blue}{Abusive} & Insulting the operator when the goal is not fulfilled. \\ \hline
 &  & Polite &  & \multicolumn{2}{l}{{\color[HTML]{9B9B9B} }} \\ \cline{3-3}
 & \multirow{-2}{*}{Positive} & Impolite & \multirow{-2}{*}{Pride, gratification} & \multicolumn{2}{l}{\multirow{-2}{*}{{\color[HTML]{9B9B9B} Not applicable to the dataset}}} \\ \cline{2-6} 
 &  & Polite &  & \textcolor{blue}{Apologetic} & Apologising for causing confusion to the operator. \\ \cline{3-3} \cline{5-6} 
\multirow{-4}{*}{User} & \multirow{-2}{*}{Negative} & Impolite & \multirow{-2}{*}{Shame, remorse, hate} & {\color[HTML]{9B9B9B} Not modelled} & {\color[HTML]{9B9B9B} Insulting the operator for no reason.} \\ \hline
 &  & Polite &  & \textcolor{blue}{Excited}, happy, anticipating & Looking forward to a good event (e.g. birthday party). \\ \cline{3-3} \cline{5-6} 
 & \multirow{-2}{*}{Positive} & Impolite & \multirow{-2}{*}{\begin{tabular}[c]{@{}l@{}}Happy-for, gloating, love,\\ satisfaction, relief, joy\end{tabular}} & \multicolumn{2}{l}{{\color[HTML]{9B9B9B} Not applicable to the dataset}} \\ \cline{2-6} 
 &  & Polite &  & \textcolor{blue}{Fearful}, sad, disappointed & Encountered a bad event (e.g. robbery). \\ \cline{3-3} \cline{5-6} 
\multirow{-4}{*}{\begin{tabular}[c]{@{}c@{}}Events,\\ facts\end{tabular}} & \multirow{-2}{*}{Negative} & Impolite & \multirow{-2}{*}{\begin{tabular}[c]{@{}l@{}}Distress, resentment, hate, fears-\\ confirmed, pity, disappointment\end{tabular}} & \multicolumn{2}{l}{\color[HTML]{9B9B9B} Not applicable to the dataset} \\ \hline
 &  & Polite &  & \textcolor{blue}{Neutral} & Describing situations and needs. \\ \cline{3-3} \cline{5-6} 
\multirow{-2}{*}{NA} & \multirow{-2}{*}{Neutral} & Impolite & \multirow{-2}{*}{NA} & {\color[HTML]{9B9B9B} Not modelled} & {\color[HTML]{9B9B9B} No emotion but rude (e.g. using imperative sentences).} \\
\bottomrule[1pt]
\end{tabular}
    \caption{Comparison between the OCC model and our labelling scheme. Emotions that do not occur in our dataset are marked as ``not applicable to our dataset''. \{User, negative, impolite\} has too few instances and \{neutral, impolite\} is not strong enough to be considered as \textit{abusive} and therefore are not modelled for now. For simplicity, emotion words in blue are used to represent each emotion category. The OCC model is illustrated in Appendix \ref{sec:the-occ-model}.}
    \label{tab:our-emotions}
\vspace*{-3mm}
\end{table*}

EmoWOZ focuses on user emotions rather than system ones. We believe recognising user emotions is the starting point for building emotion-aware task-oriented dialogue systems. We use the OCC model to arrive at specific emotion categories. For that, we consider the following aspects:

\textbf{1.\ Elicitor or cause:} The OCC model defines three main elicitors of emotion: events, agents, and objects. In task-oriented dialogues, events describe the situation which brings the user to interact with the system. For example, a user may be looking for a hotel for an upcoming trip or asking for the police information after a robbery. Agents are participants of the dialogue: the user and the system. Objects are equal to entities being talked about in the dialogue, such as the recommended hotel or the nearest police station. In our dataset, an object is always associated with either the operator, who proposes it, or an event, which drives the need for it. For this reason, we do not consider the object as an elicitor alone. On the other hand, within the agent category, it is important to distinguish between the user and the system. Therefore, we arrive at three elicitors for our annotation scheme: 1) the system, 2) the user, and 3) events (or facts).

\textbf{2.\ Valence:} In essence, the OCC model describes emotion as a valenced reaction towards an elicitor. Valence is a dimension which expresses the positivity or negativity of emotion. For example, successfully achieving a goal is likely to bring positive valence, while a misunderstanding with an agent is likely to cause negative valence. As EmoWOZ will demonstrate in a later section, valence is highly related to task success or failure, making it an important signal for a task-oriented system. We distinguish neutral and emotional utterances, and further separate emotional utterances into those with negative and positive valence.\par

\textbf{3.\ Conduct:} Conduct is not a part of the OCC model, but given the rising concern of how humans behave when interacting with virtual assistants~\cite{cercas-curry-rieser-2018-metoo}, we decided to include it. Conduct describes the politeness of users and is usually associated with emotional acts. Politeness can indicate the degree of valence. For example, the user can express very strong dissatisfaction through rudeness. It also helps distinguish emotions such as those associated with apology or abuse, which are both intrinsically negative.

Considering all combinations of these three aspects for annotation leads to a large number of classes. When choosing the final set of classes we were guided by whether or not a particular emotion category occurs in the database and the potential impact of that emotion category on the dialogue policy. We also carried out several trials and considered the ease of communicating to the annotator how to label such instances. We finally arrive at a set of 6 non-neutral emotion categories:

An emotion elicited by the operator is defined as \textit{satisfied} if it is positive, and \textit{dissatisfied} if it is negative. Positive emotion caused by an event gives us \textit{excited}, and negative \textit{fearful}. In terms of negative emotions expressed towards the system, we consider user conduct to distinguish between \textit{dissatisfied} and \textit{abusive}, since they require very different responses from the system \cite{curry2019crowd}. In terms of the negative emotions that users may direct toward themselves, we single out \textit{apologetic} behaviours since it features in human-human information-seeking dialogues. Emotion categories and their attributes in the above-mentioned aspects and their relation to the original OCC model are shown in Table \ref{tab:our-emotions}.

\subsection{Emotion Annotation Setup}
We crowd-source the emotion annotation on Amazon Mechanical Turk in a controlled manner. As suggested by \newcite{Carrin2008InfluenceOC} to improve the annotation quality, workers are shown the dialogue history up to the utterance they are required to label. Each emotion category is followed by a list of emotion words that best fit into the category and an explanation. Due to the high subjectivity in the emotion annotation \cite{Devillers2005ChallengesIR}, each dialogue is annotated by three different workers. We also implement several measures to ensure the quality of the emotion labels:\par

\paragraph{Qualification tests:} The test contains fifteen questions, seven are straight-forward and eight are more complex. The test also serves as a tutorial. For difficult questions, hints are provided to guide the workers to identify implicit emotions and use contextual information (see Appendix \ref{sec:amt-setup}).

\paragraph{Hidden tests:} We pre-label more than 1000 utterances containing obvious emotions and use them as sanity checks. The hidden tests serve as an indicator of worker reliability. If a worker scores above 80\% on the hidden tests, we assume that the worker is reliable. Otherwise, the workers' submission is subject to manual review.

\paragraph{Review for outliers:} We use a simple lexicon-based recogniser and manually annotate a small batch to have an estimate of the overall emotion distribution. If the label distribution in a worker's submissions deviates substantially from our prior belief, we mark them for manual review. 

\paragraph{Annotation limit:} We limit each worker to annotate at most 500 dialogues to ensure a diversity of workers and to avoid that workers adapt to our approval policy. Overall, we had 215 workers, each annotating 160 dialogues on average.

\section{EmoWOZ Characteristics}
\subsection{Linguistic Style}
Dialogues from MultiWOZ and DialMAGE differ linguistically. As seen in Table \ref{tab:linguistic-metrics}, DialMAGE has longer dialogues than MultiWOZ as it takes longer for the machine-generated policy to accomplish user goals. Meanwhile, users use simpler and shorter sentences when talking to a machine. Especially when the system under-performs, users are discouraged to converse with it (see sample dialogues with annotations in Appendix \ref{sec:dialogue-examples}). We will analyse the impact of these differences on emotion recognition in Section \ref{sec:results-and-discussion}.

\begin{table}[!htbp]
\setlength\tabcolsep{1.2pt}
\footnotesize
\centering
\begin{tabular}{l|rrr}
\toprule[1pt]
 & \textbf{MultiWOZ} & \textbf{DialMAGE} & \textbf{EmoWOZ}\\\hline
\# Dialogues & 10,438 & 996 & 11,438\\
\# Unique tokens & 27,833 & 3,133 & 28,417 \\
Avg. turns / dialogue & 13.7 & 24.3 & 14.6 \\
Avg. tokens / user turn & 11.6 & 5.7 & 10.6 \\
\begin{tabular}[c]{@{}c@{}}Avg. unique user \\ tokens / dialogue\end{tabular}\ & 57.8 & 36.5 & 55.6\\
\bottomrule[1pt]
\end{tabular}
\caption{Comparison of linguistic features in EmoWOZ and its subsets.}
\label{tab:linguistic-metrics}
\vspace*{-3mm}
\end{table}

\subsection{Emotion Distribution}

According to Table \ref{tab:emo-distribution}, the most common non-neutral emotion in EmoWOZ is \textit{satisfied}, followed by \textit{dissatisfied}. This is expected in task-oriented dialogues as users mainly express emotion in relation to their goals. While MultiWOZ contains more neutral utterances, it has a more diverse emotion distribution than DialMAGE. MultiWOZ contributes most \textit{satisfied} utterances whereas DialMAGE contributes most \textit{dissatisfied} utterances. This is in line with their respective dialogue-generating setup.

Sometimes users also express emotion to engage or provoke the operator. MultiWOZ contains more \textit{apologetic} and less \textit{abusive} utterances than DialMAGE, suggesting that users tend to be more polite when talking to human operators. Dialogues from MultiWOZ also contain more event-elicited emotions (\textit{fearful} and \textit{excited}) than DialMAGE. Users are more talkative when conversing with human operators. Users may describe a miserable situation they were experiencing, hoping to be helped and comforted. A human operator would naturally show empathy. In MultiWOZ, the operator sometimes asks if the user is alright when the user is looking for help from a robbery. When talking to machines, users tend not to express such chit-chat-style emotions due to the expected incapability of the machine to reciprocate. This indicates that an emotionally intelligent agent will allow dialogues that are emotionally richer and more nuanced, even in a task-oriented setting. \par

\begin{table}[H]
\centering
\footnotesize
\setlength\tabcolsep{1.4pt}
\begin{tabular}{l|rr|rr|rr}
\toprule[1pt]
\multirow{2}{*}{\textbf{Emotion}} & \multicolumn{2}{c|}{\textbf{EmoWOZ}} & \multicolumn{2}{c|}{\textbf{MultiWOZ}} & \multicolumn{2}{c}{\textbf{DialMAGE}} \\
 & Count & Prop. & Count & Prop. & Count & Prop. \\\hline
Neutral & 58,656 & 70.1\% & 51,426 & 71.9\% & 7,230 & 59.8\% \\
Fearful & 396 & 0.5\% & 381 & 0.5\% & 15 & 0.1\% \\
Dissatisfied & 5,117 & 6.1\% & 914 & 1.3\% & 4,203 & 34.8\% \\
Apologetic & 840 & 1.0\% & 838 & 1.2\% & 2 & 0.02\% \\
Abusive & 105 & 0.2\% & 44 & 0.1\% & 61 & 0.5\% \\
Excited & 971 & 1.2\% & 860 & 1.2\% & 111 & 0.9\% \\
Satisfied & 17,532 & 21.0\% & 17,061 & 23.8\% & 471 & 3.9\%\\
\bottomrule[1pt]
\end{tabular}
\caption{Count and prop(ortion) of emotion labels.}
\label{tab:emo-distribution}
\vspace*{-3mm}
\end{table}

\begin{table*}[t]
\centering
\footnotesize
\setlength\tabcolsep{2.6pt}
\begin{tabular}{ccc|lllllll|ll|ll|ll}
\toprule[1pt]
 
\multirow{2}{*}{Model} & \multirow{2}{*}{Feature} & \multirow{2}{*}{Ctx.} & \multicolumn{7}{c|}{\textbf{F1 of Each Emotion in EmoWOZ}} & \multicolumn{2}{c|}{\textbf{EmoWOZ}} & \multicolumn{2}{c|}{\textbf{MultiWOZ}} & \multicolumn{2}{c}{\textbf{DialMAGE}} \\
 &  &  & \multicolumn{1}{c}{Neu.} & \multicolumn{1}{c}{Fea.} & \multicolumn{1}{c}{Dis.} & \multicolumn{1}{c}{Apo.} & \multicolumn{1}{c}{Abu.} & \multicolumn{1}{c}{Exc.} & \multicolumn{1}{c|}{Sat.} & \multicolumn{1}{c}{Mac.} & \multicolumn{1}{c|}{Wgt.} & \multicolumn{1}{c}{Mac.} & \multicolumn{1}{c|}{Wgt.} & \multicolumn{1}{c}{Mac.} & \multicolumn{1}{c}{Wgt.} \\ \hline
BERT & BERT & No & 89.8 & 36.2 & 35.1 & 70.4 & 27.5 & 42.9 & 88.8 & 50.1 & 73.5 & \textbf{48.4} & \textbf{83.2} & 42.7 & 43.8 \\
ContextBERT & BERT & Yes & \textbf{92.1}* & 30.1 & \textbf{61.7}* & 62.4 & \textbf{41.7} & 40.8 & \textbf{89.1} & 54.3 & \textbf{79.7}* & 45.1 & 83.1 & 50.0 & \textbf{73.5}* \\
DialogueRNN & GloVe & Yes & 83.5 & 12.7 & 51.4 & 57.7 & 0.0 & 32.7 & 86.4 & 40.1 & 74.6 & 34.1 & 79.2 & 43.2 & 61.2 \\
DialogueRNN & BERT & Yes & 86.9 & 41.3 & 47.5 & \textbf{71.5} & 25.6 & 39.4 & 87.6 & 52.1 & 75.5 & 44.5 & 81.9 & 51.4 & 60.6 \\
COSMIC & BERT+COMET & Yes & 89.8 & \textbf{52.0}* & 50.7 & 70.9 & 31.6 & \textbf{44.4} & 88.4 & \textbf{56.3} & 77.1 & 46.7 & 82.7 & \textbf{57.2} & 61.7 \\
\bottomrule[1pt]
\end{tabular}
\caption{Comparison of baseline models. We report the F1 for each emotion label (\textbf{Neu}tral, \textbf{Fea}rful, \textbf{Dis}satisfied, \textbf{Apo}logetic, \textbf{Abu}sive, \textbf{Exc}ited, \textbf{Sat}isfied) on EmoWOZ as well as \textbf{Mac}ro and \textbf{W}ei\textbf{g}h\textbf{t}ed F1 (excluding neutral) on EmoWOZ and its subsets. ``Ctx.'' stands for ``context''. * indicates statistically significant difference with $p<0.05$ between the best and the second best values in each column. Please refer to Appendix \ref{sec:emotion-classification-results} for more detailed results. 
\label{tab:f1-scores}}
\vspace*{-3mm}
\end{table*}

\subsection{Inter-annotator Agreement}
We measure the inter-annotator agreement by computing Fleiss' Kappa~\cite{fleiss1971measuring}. Fleiss' Kappa for EmoWOZ is 0.602, suggesting a substantial agreement. Fleiss' Kappa for MultiWOZ is 0.611, higher than 0.465 for DialMAGE. Emotions in DialMAGE are more challenging to annotate because users express emotion less explicitly when they know that they are talking to a machine that does not react to emotions. Annotators often have to infer the user's implicit emotions from dialogue history, for example, based on repetitions or misunderstanding.\par

\begin{figure}[!htbp]
\centering
\includegraphics[width=0.4\textwidth]{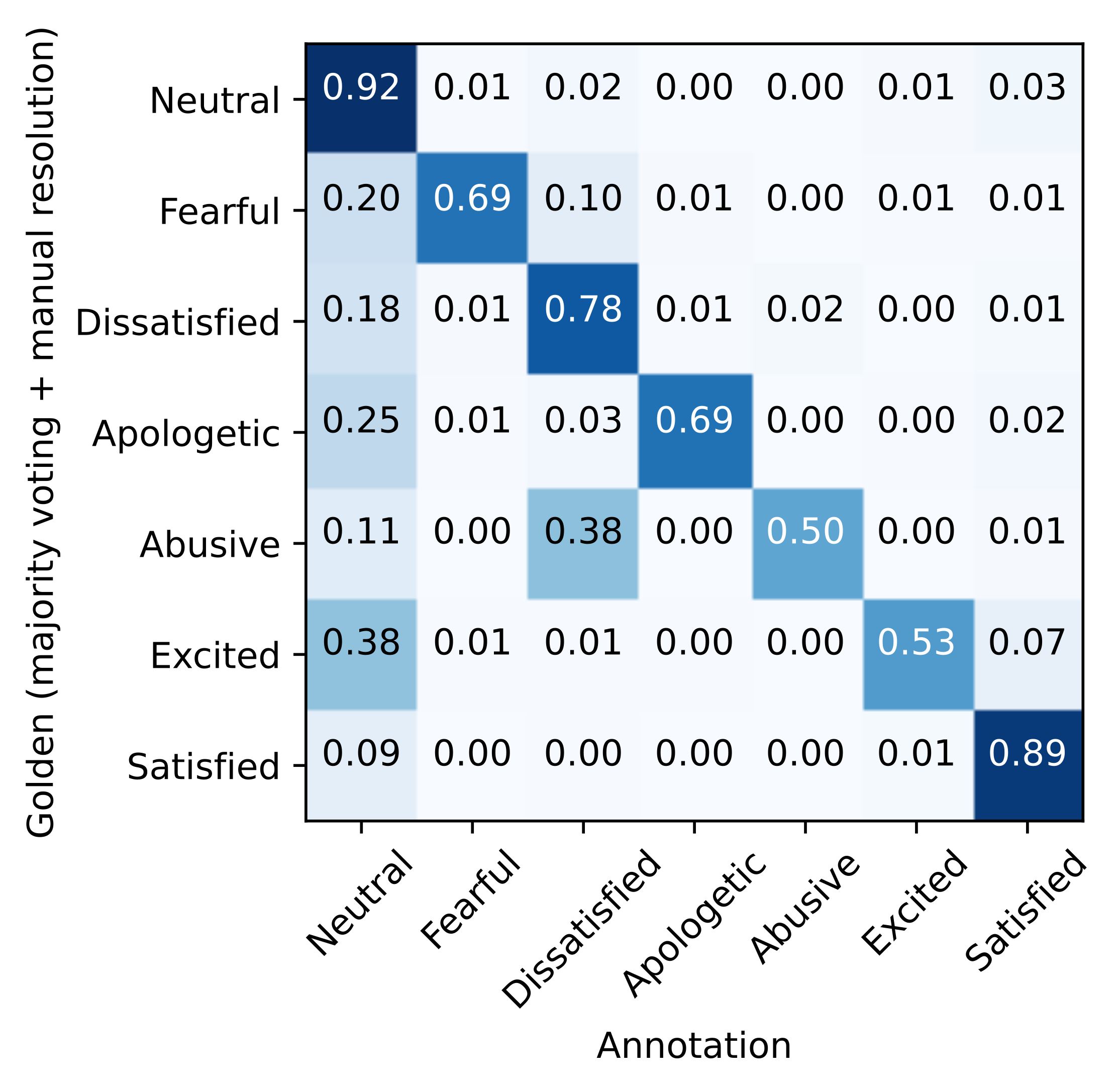}
\vspace*{-3mm}
\caption{Confusion matrix of emotion annotations.}
\label{fig:annotator-cm}
\vspace*{-3mm}
\end{figure}
    
Among all utterances, 72.1\% see a full agreement among three annotators, 26.4\% see a partial agreement, and 1.5\% see no agreement. The count of each case in each subset can be found in Appendix \ref{sec:annotator-agreement}. Utterances for which no agreement is reached are resolved manually.

Figure \ref{fig:annotator-cm} illustrates the confusion matrix between annotators' labels and the golden labels. Most disagreements occur between non-neutral emotions and neutral, as well as \textit{abusive} and \textit{dissatisfied}. A reasonable explanation is that workers adopt different valence or impoliteness thresholds when they make decisions. Note that dissatisfied is rarely confused with abusive, but rather with neutral, suggesting that the ambiguity lies in when an expression of dissatisfaction is considered to be rude or abusive, and not due to the similarity between abuse and dissatisfaction.

On the other hand, confusions between \textit{fearful} and \textit{dissatisfied} suggest workers may also interpret elicitors differently.
For example, a user may express negative emotions after the agent informed that there is no attraction meeting the user's criteria. While the emotion is caused by the fact that there is no match, one can also argue that the operator failed to suggest alternative options. We believe differences on interpretations are natural to a certain extent, as emotion appraisal may differ across individuals \cite{kuppens2007individual}. 

\section{Experiment}
\subsection{Emotion Recognition in Dialogue}

Emotion recognition aims to recognise emotion within an utterance. Unlike utterances in isolation, emotion recognition in dialogues is highly contextual with respect to the dialogue history. As baselines, we compare two models originally developed for chit-chat emotion recognition as well as various BERT-based models. We believe emotion recognition is the first step towards an emotion-aware task-oriented agent, as a means for a deployed agent to obtain emotion information during an interaction.

\subsubsection{Baselines}

\textbf{BERT~\cite{devlin-etal-2019-bert}:} BERT is used as the utterance encoder. Each user turn is encoded in isolation without any dialogue context. The \verb|[CLS]| token from a \verb|bert-base-cased| model is used as the feature representation, which is then fed into a linear output layer for classification.

\textbf{ContextBERT:} The set-up is identical to that of BERT, except that the entire dialogue history and the current user utterance are concatenated in the reversed order to form one long sequence. We add ``\verb|User:|'' and ``\verb|System:|'' to mark the speaker of each turn. 

\textbf{DialogueRNN~\cite{Majumder_Poria_Hazarika_Mihalcea_Gelbukh_Cambria_2019}:} The model combines gated recurrent units (GRUs) with an attention mechanism to capture the long-term trajectory of the dialogue. We experiment with using GloVe embeddings~\cite{pennington-etal-2014-glove} or the \verb|[CLS]| representation from BERT as input features. When GloVe is used, a convolutional neural network (CNN) layer is used as a feature extractor to generate utterance representations. This CNN layer is dropped when using BERT features.

\textbf{COSMIC~\cite{ghosal-etal-2020-cosmic}:} This model also combines GRUs with the attention mechanism. In addition to utterance representations from a pre-trained language model (LM), it supplements input features with common-sense knowledge extracted from a pre-trained commonsense transformer model called COMET~\cite{bosselut-etal-2019-comet}. Although the original paper uses RoBERTa as input features, we found that BERT results in a better sequence representation for emotion recognition on our data. Therefore we use BERT as the utterance encoder in our experiments.

\begin{figure*}
    \centering
    \includegraphics[width=\textwidth]{{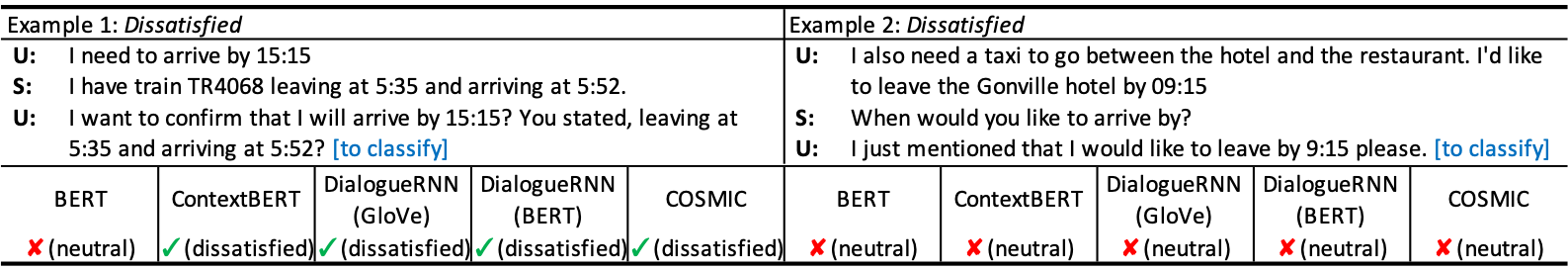}}
    \caption{Example dialogues and the emotion prediction for the last utterance by each model.}
    \label{fig:example-predictions}
\end{figure*}

\begin{table*}[t]
\centering
\footnotesize
\setlength\tabcolsep{2.4pt}
\begin{tabular}{l|lllllllll|lllllllll}
\toprule[1pt]
\multirow{2}{*}{\begin{tabular}[c]{@{}l@{}}Training\\ Data\end{tabular}} & \multicolumn{9}{c|}{\textbf{Test on MultiWOZ}} & \multicolumn{9}{c}{\textbf{Test on DialMAGE}} \\ 
 & \multicolumn{1}{c}{Neu.} & \multicolumn{1}{c}{Fea.} & \multicolumn{1}{c}{Dis.} & \multicolumn{1}{c}{Apo.} & \multicolumn{1}{c}{Abu.} & \multicolumn{1}{c}{Exc.} & \multicolumn{1}{c}{Sat.} & \multicolumn{1}{c}{Mac.} & \multicolumn{1}{c|}{Wgt.} & \multicolumn{1}{c}{Neu.} & \multicolumn{1}{c}{Fea.} & \multicolumn{1}{c}{Dis.} & \multicolumn{1}{c}{Apo.} & \multicolumn{1}{c}{Abu.} & \multicolumn{1}{c}{Exc.} & \multicolumn{1}{c}{Sat.} & \multicolumn{1}{c}{Mac.} & \multicolumn{1}{c}{Wgt.} \\ \hline
MultiWOZ & \textbf{95.1}* & \textbf{35.7} & \textbf{36.4}* & \textbf{70.3}* & \textbf{19.4} & 34.1 & \textbf{90.0} & \textbf{47.7} & \textbf{83.9} & 80.2 & \textbf{11.7} & 7.7 & \textbf{43.7} & 11.9 & \textbf{60.1} & 66.3 & 33.6 & 14.5 \\
DialMAGE & 89.4 & 0 & 11.2 & 0 & 0 & 13.9 & 77.3 & 17.0 & 67.8 & 72.1 & 0 & \textbf{75.7} & 0 & 5.0 & 58.6 & \textbf{71.7} & 35.2 & 72.9 \\
EmoWOZ & 93.5 & 33.7 & 30.4 & 62.4 & 17.3 & \textbf{37.1} & 89.8 & 45.1 & 83.1 & \textbf{81.6} & 5.0 & 75.5 & 40.0 & \textbf{52.8}* & 57.3 & 69.2 & \textbf{50.0}* & \textbf{73.5} \\
\bottomrule[1pt]
\end{tabular}
\caption{Performance of ContextBERT in cross-dataset experiments. We report the F1 for each emotion label (\textbf{Neu}tral, \textbf{Fea}rful, \textbf{Dis}satisfied, \textbf{Apo}logetic, \textbf{Abu}sive, \textbf{Exc}ited, \textbf{Sat}isfied), as well as \textbf{Mac}ro and \textbf{W}ei\textbf{g}h\textbf{t}ed F1 (excluding neutral). * indicates statistically significant difference with $p<0.05$ between the best and the second best values in each column. For detailed results, please refer to Appendix \ref{sec:emotion-classification-results}.  \label{tab:f1-cross-emotion-short}}
\vspace*{-3mm}
\end{table*}

\subsubsection{Experimental Setup}
We perform a recognition task on the 7 emotions proposed in our annotation scheme\footnote{We also performed the same experiments on 3 sentiment labels. Results can be found in Appendix \ref{sec:sentiment-classification-results}.}.
All models are implemented in PyTorch~\cite{NEURIPS2019_9015}.
For COSMIC and DialogueRNN, we use the code provided by the respective papers. 
We include more details on the hyperparameters of each model in Appendix \ref{sec:hyperparameters}.
To split EmoWOZ into training, validation, and testing sets, we keep the original split of MultiWOZ and further divide DialMAGE with a ratio of 8:1:1, leading to 9,234, 1,100, and 1,100 dialogues in each set. We run each task on 5 different seeds and report the average performance.

\subsubsection{Results and Discussion}
\label{sec:results-and-discussion}
\paragraph{Recognition on emotion classes.} Table \ref{tab:f1-scores} summarises the performance of baseline models. Since almost 70\% of the annotations are \textit{neutral}, we exclude it when calculating average F1 scores. In general, models that take into account context information perform better on the full EmoWOZ. This shows the importance of context or dialogue-level features in emotion recognition in task-oriented dialogues. An exception is DialogueRNN with GloVe feature, which underperforms in EmoWOZ macro F1, likely due to the non-contextual embedding used. On the other hand, BERT scores very well on MultiWOZ dialogues but performs poorly on DialMAGE for both setups. This suggests that emotions in MultiWOZ are less context-dependent.\par

BERT, the only non-contextual model among our baselines, performs well for \textit{apologetic}, \textit{excited}, and \textit{satisfied}, potentially due to the existence of distinguishable keywords associated with these emotions such as ``thank you'' for \textit{satisfied} and ``sorry'' for \textit{apologetic}. These emotion labels do not benefit much from context. 
In contrast, BERT produces a significantly worse F1 on \textit{dissatisfied}, probably because users tend to express dissatisfaction more implicitly, for instance via repetition or correction, making dialogue-level features necessary.

Figure \ref{fig:example-predictions} shows two dialogues with implicit emotions and predictions made by respective baseline models. In example 1, the system gives the wrong time of arrival, eliciting mild annoyance from the user. BERT predicts \textit{neutral} because in isolation, the utterance has no words suggesting dissatisfaction. All other models correctly recognise \textit{dissatisfied}, as they capture the misunderstanding occurs in previous dialogue turns. Example 2 presents a similar but more implicit case, where all models fail. This shows that EmoWOZ contains contextualised emotions that are more implicit and subtle, requiring more sophisticated features and models.

\paragraph{Complementarity between MultiWOZ and DialMAGE.} 
Due to different linguistic features and emotion distributions in MultiWOZ and DialMAGE, one concern is that the models learn to predict emotion based on these statistical artifacts. According to Table \ref{tab:linguistic-metrics}, the most obvious difference is the average utterance length (5.8 in DialMAGE and 11.8 in MultiWOZ). A naive model may simply recognise the data source from word count and predict the most likely emotion from that source. Table \ref{tab:artifact} presents how ContextBERT trained on EmoWOZ predicts emotion in long DialMAGE and short MultiWOZ utterances. The emotion distribution in model prediction is vastly different from that in the complementing subset. Clearly, the model does not simply count words to decide on the underlying emotion.

\begin{table}[ht]
\centering
\scriptsize
\begin{tabular}{l|rr}
\toprule[1pt]
    & \textbf{Dissatisfied} & \textbf{Satisfied} \\ \hline
MultiWOZ Label          & 1.5\%       & 24.0\%   \\
DialMAGE (\#token$>$11.8) Label                   & 28.8\%      & 1.2\%    \\
DialMAGE (\#token$>$11.8) Prediction  & 35.5\%      & 1.5\%    \\ \hline
DialMAGE Label           & 39.3\%      & 4.0\%    \\ 
MultiWOZ (\#token$<$5.8) Label                  & 1.2\%       & 37.7\%   \\
MultiWOZ (\#token$<$5.8) Prediction & 3.0\%       & 38.9\%   \\
\bottomrule[1pt]
\end{tabular}
\caption{Emotion distribution in labels and ContextBERT prediction. See Appendix \ref{sec:emo-dist-model-pred} for full results. 
\label{tab:artifact}}
\vspace*{-3mm}
\end{table}

Table \ref{tab:f1-cross-emotion-short} presents cross-data experiments with ContextBERT, examining how well the two subsets complement each other. Complementing DialMAGE with dialogues from MultiWOZ improves the macro F1 and the F1 score of \textit{abusive} significantly. On the other hand, while complementing MultiWOZ with DialMAGE leads to a slight improvement in the F1 score of \textit{excited}, other F1 scores decrease to various extent.\par

\paragraph{Recall and precision on satisfied and dissatisfied for task-oriented dialogue.} We further investigate the change in F1 of each emotion on MultiWOZ by looking at the change in recall and precision after complementing MultiWOZ with DialMAGE. We believe it is necessary to distinguish recall and precision, as for some emotions, one may be more important than the other. The relative importance of recall and precision for each emotion class depends on its implication to a task-oriented dialogue system and the consequence of false recognition. Most importantly for task-oriented dialogue system, a high recall of \textit{dissatisfied} is desirable because the system should not miss any failure in dialogues. Failing to recognise dissatisfaction can trigger more anger from the user and therefore impair task completion (see Figure \ref{fig:annotation-examples}). On the other hand, a high precision may be more desirable for all other emotions to ensure proper affective response from the system. When the relative importance of recall and precision of the emotion is taken into account, complementing MultiWOZ with DialMAGE is beneficial to \{\textit{dissatisfied}\} for higher recall and \{\textit{fearful}, \textit{excited}, \textit{satisfied}\} for higher precision, see Table~\ref{tab:precision-recall-short}. Detailed results can be found in Appendix \ref{sec:precision-recall}.

\begin{table}[H]
\centering
\small
\setlength\tabcolsep{2pt}
\begin{tabular}{l|llllll}
\toprule[1pt]
\textbf{Metric} & \multicolumn{1}{c}{\textbf{Fea.}} & \multicolumn{1}{c}{\textbf{Dis.}} & \multicolumn{1}{c}{\textbf{Apo.}} & \multicolumn{1}{c}{\textbf{Abu.}} & \multicolumn{1}{c}{\textbf{Exc.}} & \multicolumn{1}{c}{\textbf{Sat.}} \\ \hline
Recall & \textminus 6.7** & +29.1** & \textminus 7.8 & +8.0 & +0.6 & -0.4\\
Precision & +7.5* & \textminus 22.8** & \textminus 7.9 & \textminus 11.3 & +7.1** & +0.1\\
\bottomrule[1pt]
\end{tabular}
\caption{Change in precision and recall for each emotion label (\textbf{Fea}rful, \textbf{Dis}satisfied, \textbf{Apo}logetic, \textbf{Abu}sive, \textbf{Exc}ited, \textbf{Sat}isfied) on MultiWOZ by ContextBERT, after adding DialMAGE to training. ** and * indicate statistically significant changes with $p<0.05$ and $p<0.1$ respectively. \label{tab:precision-recall-short}}
\vspace*{-3mm}
\end{table}

\subsection{Emotions for Dialogue State Tracking}
In task-oriented dialogues, dialogue state tracking (DST) aims to continuously track the user's goal and intent as the dialogue progresses~\cite{YOUNG2010150}. We hypothesise that the user emotion can help inform the system about their goal. To investigate this, we train a dialogue state tracker that incorporates an additional task to predict one of 7 emotional classes on MultiWOZ 2.1~\cite{eric-etal-2020-multiwoz}. We utilise the \textit{out-of-task training} approach and the available code presented in~\cite{heck-etal-2020-task}. We follow the multitask learning (MTL) algorithm, where on each training step, the same model is trained on two different batches, one from the main task (DST) and one from the auxiliary task (emotion recognition). Since neutral emotion provides limited information on the user goal, we remove a half of the neutral utterances when performing MTL. We show that additional emotion labels can lead to a significant improvement ($p < 0.02$) in the joint goal accuracy (JGA) of DST (see Table~\ref{tab:mtl}).

\begin{table}[!htbp]
    \centering
    \small
    \setlength\tabcolsep{5pt}
    \begin{tabular}{l|c}
        \toprule[1pt]
        \textbf{Training tasks}  & \textbf{JGA} \\ \hline
        Dialogue state tracking  & 53.7 \\
        Dialogue state tracking \& emotion recognition & \textbf{54.7} \\
        \bottomrule[1pt]
    \end{tabular}
    \caption{JGA of DST on MultiWOZ.}
    \label{tab:mtl}
\vspace*{-3mm}
\end{table}

\section{Conclusion}

In this work, we examined emotions and their expression in the context of task-oriented dialogues, where emotions are centred around a user goal. We used the OCC model as a starting point to derive a comprehensive annotation scheme beyond sentiment polarity for emotions in relation to user goals. We designed a set of 7 emotions that differ in terms of valence, conduct and elicitor to capture the cognitive context of emotions, while maintaining labeling simplicity. 

With EmoWOZ, we present a publicly available, large-scale human-annotated emotion corpus consisting of Wizard-of-Oz style as well as dialogues with a machine-generated policy. Our intention with EmoWOZ is to overcome the lack of large emotion-labelled corpora to support research towards emotion-aware task-oriented dialogue systems, for dialogues closer to human-human interactions.

We apply various emotion recognition models to EmoWOZ and examined the effect of context for different emotions. In cross-dataset experiments we analysed the complementarity of WOZ-style data and machine-generated policy data. Our results show that recognising context-dependent and implicit emotions from task-oriented dialogues is a challenging task that will benefit from further research. EmoWOZ provides an ideal test bed for that. Lastly, we leveraged emotion recognition in the dialogue state tracking task to exemplify the utility of emotion labels in dialogue modeling. 

We hope this dataset can offer insights beyond the scope of emotion recognition and push the performance of downstream tasks in task-oriented dialogue modelling. In future work, we plan to investigate tailored models for emotion recognition in task-oriented dialogues that take advantage of high-level features such as dialogue acts or belief states. We are also interested in using emotion as a feedback signal within reinforcement learning policy optimisation.

\section{Acknowledgements}
S. Feng, N. Lubis, M. Heck, and C. van Niekerk are supported by funding provided by the Alexander von Humboldt Foundation in the framework of the Sofja Kovalevskaja Award endowed by the Federal Ministry of Education and Research, while C. Geishauser and H-C. Lin are supported by funds from the European Research Council (ERC) provided under the Horizon 2020 research and innovation programme (Grant agreement No. STG2018 804636). Computing resources were provided by Google Cloud.
\newpage
\section{Bibliographical References}\label{reference}

\bibliographystyle{lrec2022-bib}
\bibliography{lrec2022-example}


\newpage
\appendix
\onecolumn
\counterwithin{figure}{section}
\section{The OCC Model}
\label{sec:the-occ-model}
\begin{figure}[!htbp]
    \centering
    \includegraphics[scale=0.75]{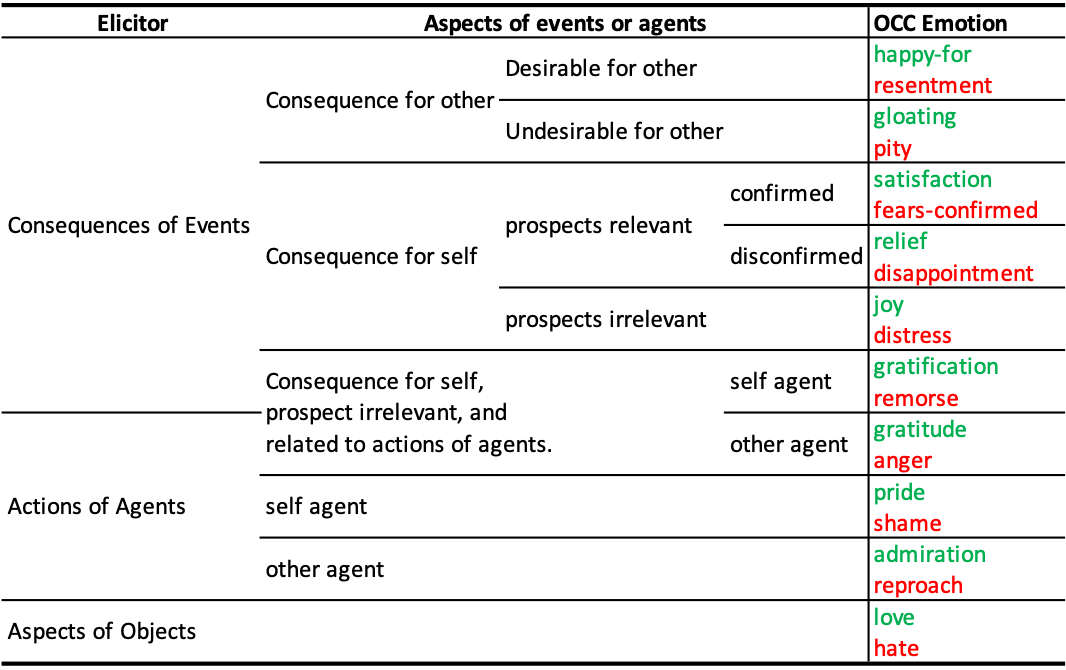}
    \caption{The OCC Model}
    \label{fig:occ-model}
\end{figure}

Figure \ref{fig:occ-model} summarises definitions of emotion groups in the OCC model.

\newpage
\counterwithin{figure}{section}
\section{Amazon Mechanical Turk Set-up}
\label{sec:amt-setup}

\subsection{Qualification Test}
\begin{figure}[!htbp]
    \centering
    \includegraphics[width=\textwidth]{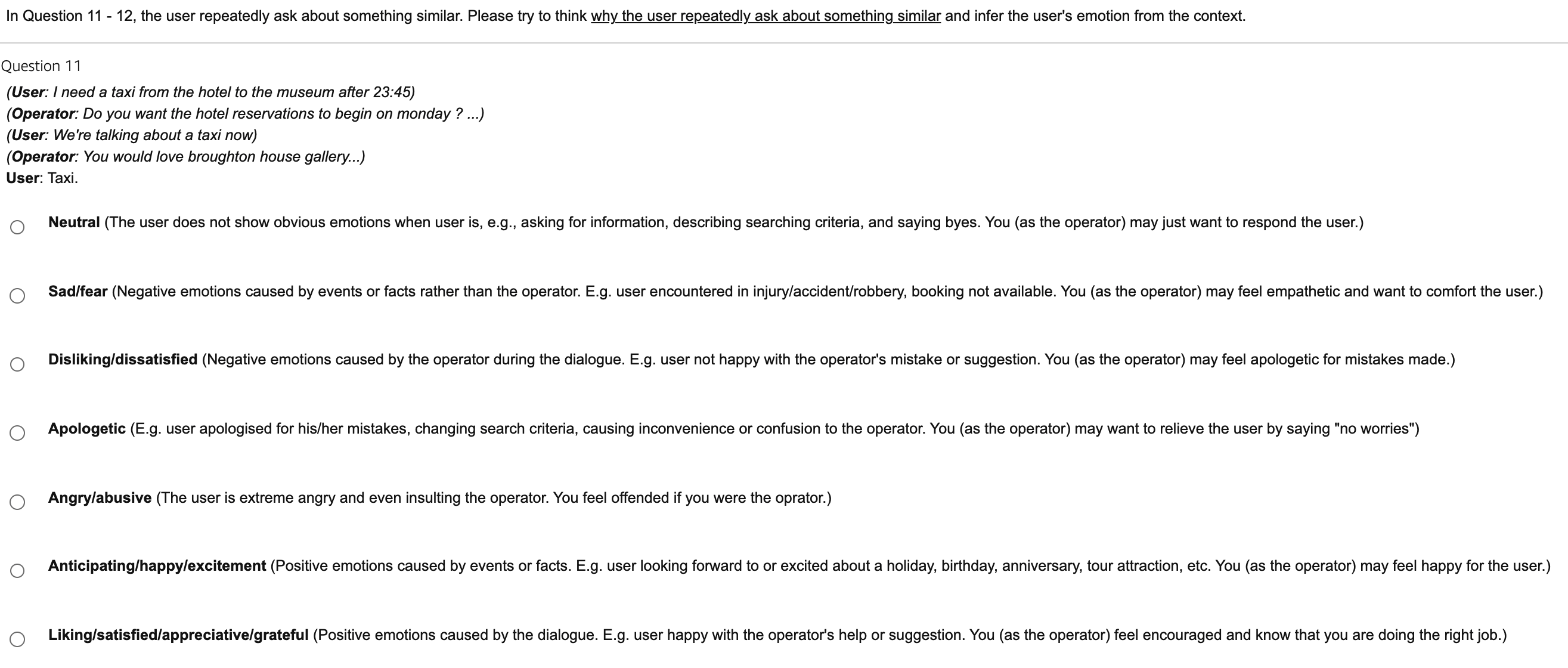}
    \caption{One of fifteen questions in our qualification test}
    \label{fig:qualification-test}
\end{figure}

Figure \ref{fig:qualification-test} illustrates one example from our qualification test. Hints are provided for difficult questions containing implicit emotions as shown in the example. 

\subsection{Main Task Page}
\begin{figure}[!htbp]
    \centering
    \includegraphics[width=\textwidth]{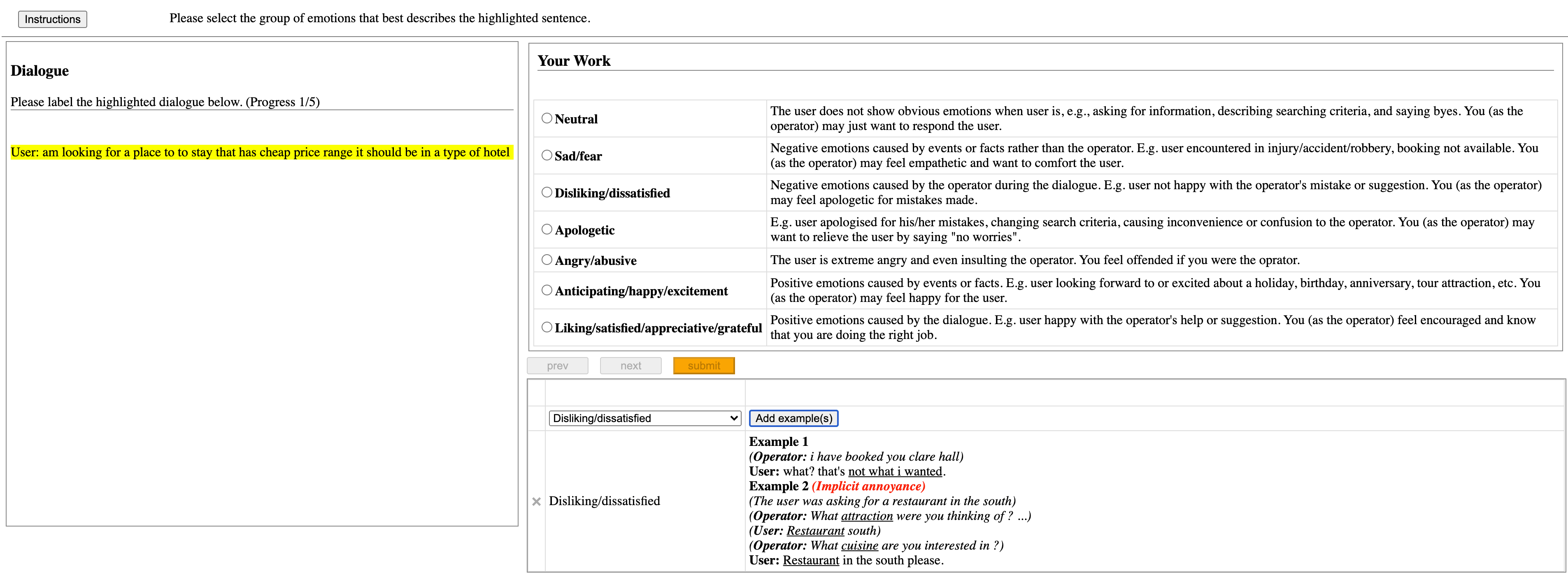}
    \caption{Amazon Mechanical Turk main task page}
    \label{fig:taskpage}
\end{figure}

Figure \ref{fig:taskpage} shows the task page for workers. Before arriving at this page, they will be prompted with a consent form and a message asking if they would like to go through a tutorial.

\newpage
\counterwithin{figure}{section}
\section{Dialogue Examples}
\label{sec:dialogue-examples}
\begin{figure}[!htbp]
    \centering
    \includegraphics[scale=0.65]{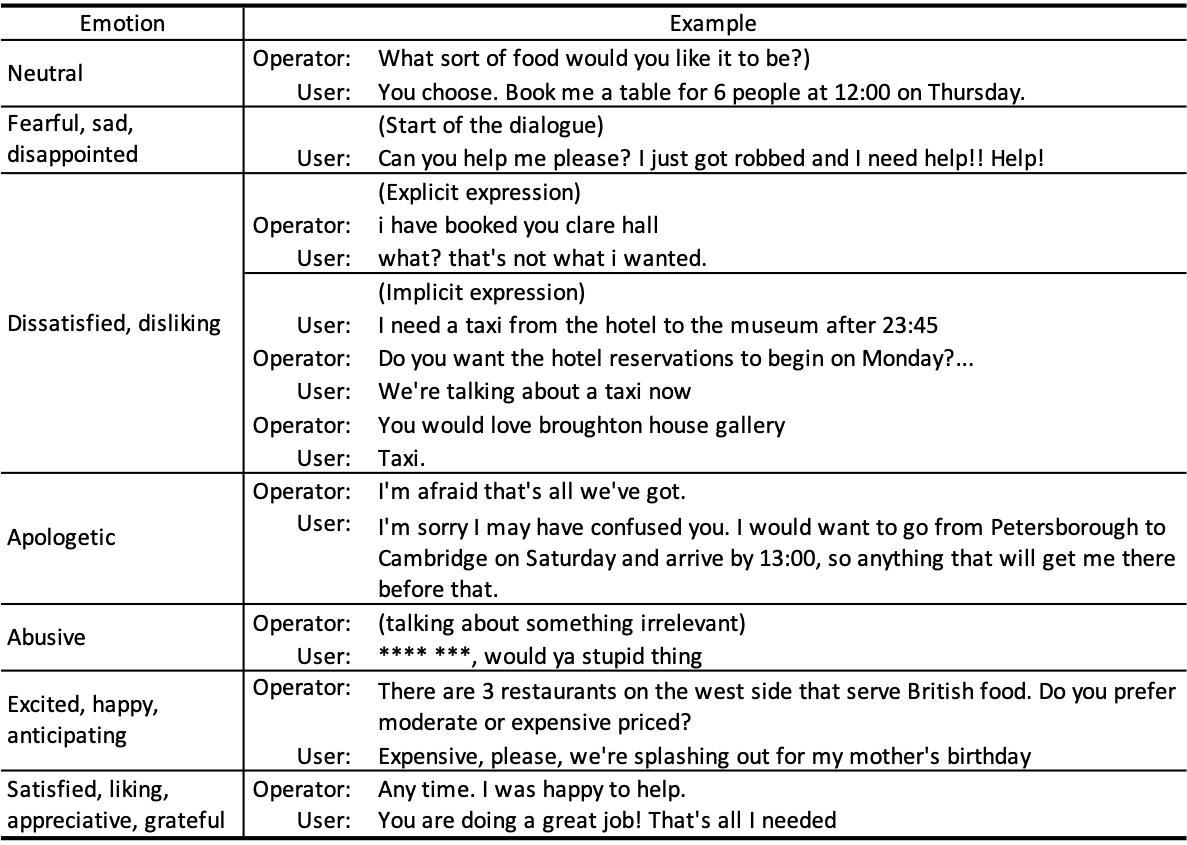}
    \caption{Example for each emotion label}
    \label{fig:emotion-examples}
\end{figure}

\begin{figure}[!htbp]
    \centering
    \includegraphics[scale=0.7]{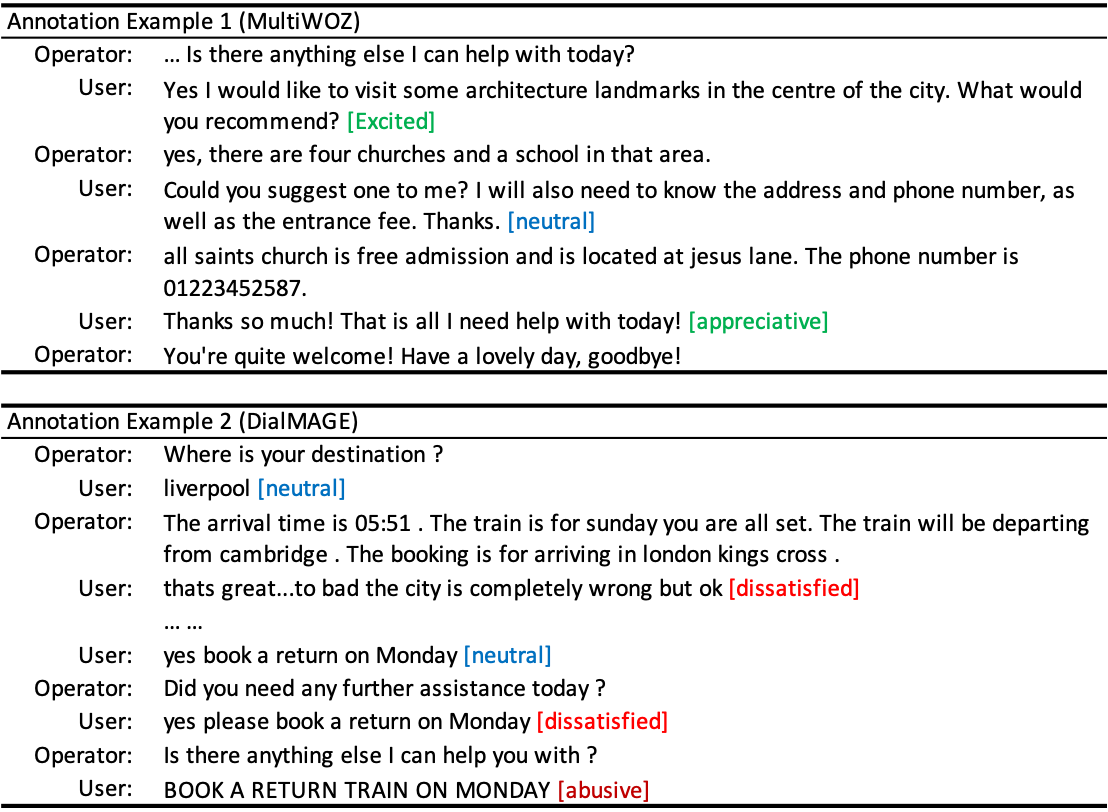}
    \caption{Annotation examples from EmoWOZ}
    \label{fig:annotation-examples}
\end{figure}

Figure \ref{fig:emotion-examples} shows examples of how emotions are expressed by the user in EmoWOZ. Figure \ref{fig:annotation-examples} shows examples of annotated dialogues.

\newpage
\setcounter{table}{0}
\renewcommand{\thetable}{\Alph{section}\arabic{table}}
\section{Annotator Agreement in EmoWOZ}
\label{sec:annotator-agreement}
\begin{table}[!htbp]
\centering
\small
\setlength\tabcolsep{3pt}
\begin{tabular}{ccccc}
\toprule[1pt]
\textbf{Source} & \textbf{Fleiss' Kappa} & \textbf{\#NA} & \textbf{\#PA} & \textbf{\#FA}  \\
\hline
MultiWOZ & 0.611 & 1,016 & 17,270 & 53,238  \\
DialMAGE & 0.465 & 259 & 4,773 & 7,061 \\
\hline
EmoWOZ (overall) & 0.602 & 1,275 & 22,043 & 60,299\\
\bottomrule[1pt]
\end{tabular}
\caption{Inter-annotator agreement and agreement count of EmoWOZ and its subsets. NA means no agreement---three annotators annotate with three different emotions. PA means partial agreement---only two annotators annotate with the same emotion. FA means full agreement---three annotators annotate with the same emotion.}
\label{tab:agreement}
\end{table}

\setcounter{table}{0}
\renewcommand{\thetable}{\Alph{section}\arabic{table}}
\section{Hyperparameters for Model Training}
\label{sec:hyperparameters}
\begin{table}[!htbp]
    \centering
    \begin{tabular}{c|c|c|c|c}
    \toprule[1pt]
        \textbf{Model} & \textbf{Optimiser} & \textbf{Learning Rate} & \textbf{L2 Reguliser Weight} & \textbf{Training Epochs} \\
        \hline
        BERT & Adam & 2e-5 & 0 & 10 \\
        ContextBERT & Adam & 2e-5 & 0 & 10 \\
        DialogueRNN(GloVe) & Adam & 1e-4 & 1e-5 & 60 \\
        DialogueRNN(BERT) & Adam & 1e-4 & 1e-4 & 60 \\
        COSMIC & Adam & 1e-4 & 3e-4 & 20 \\
        \bottomrule[1pt]
    \end{tabular}
    \caption{Hyperparameters for model training}
    \label{tab:hyperparameters}
\end{table}

\newpage
\setcounter{table}{0}
\renewcommand{\thetable}{\Alph{section}\arabic{table}}
\section{Detailed Cross-dataset Experiment Results}
\label{sec:f1-scores}

\subsection{Emotion Classification (7 classes)}
\label{sec:emotion-classification-results}
\begin{table}[ht!]
\centering
\scriptsize
\setlength\tabcolsep{3pt}

\begin{tabular}{l|c|ccccccc|ccc|ccc}
\toprule[1pt]
\multirow{2}{*}{\textbf{Model}} & \multicolumn{1}{c|}{\multirow{2}{*}{\textbf{Set-up}}} & \multicolumn{7}{c|}{\textbf{F1 for each Emotion Label}} & \multicolumn{3}{c|}{\textbf{Average F1 w/o Neutral}} & \multicolumn{3}{l}{\textbf{Average F1 w Neutral}} \\
 &  & Neutral & Fearful & Dissatisfied & Apologetic & Abusive & Excited & Satisfied & Micro & Macro & Weighted & Micro & Macro & Weighted \\ \hline
\multirow{7}{*}{BERT} & D$\,\to\,$D & 59.75 & 0 & \textbf{50.34} & 0 & 12.99 & 61.42 & \textbf{72.43} & \textbf{52.50} & 32.86 & \textbf{51.45} & 59.08 & 36.70 & 55.94 \\
 & M$\,\to\,$D & \textbf{71.57} & \textbf{11.67} & 1.36 & \textbf{100} & 6.15 & \textbf{64.30} & 68.85 & 16.97 & 42.05 & 9.36 & 56.94 & 46.27 & 43.02 \\
 & E$\,\to\,$D & 69.94 & 0 & 41.43 & 60.0 & \textbf{29.41} & 56.01 & 69.13 & 45.47 & \textbf{42.66} & 43.82 & \textbf{61.09} & \textbf{46.56} & \textbf{57.95} \\ \cline{2-15}
  & D$\,\to\,$M & 71.09 & 0 & 6.02 & 0 & 11.11 & 15.60 & 88.07 & 62.63 & 20.13 & 77.16 & 70.58 & 27.41 & 72.77 \\
 & M$\,\to\,$M & \textbf{95.34} & \textbf{43.00} & \textbf{40.87} & \textbf{73.03} & 19.05 & \textbf{40.45} & \textbf{90.39} & \textbf{85.19} & \textbf{51.13} & \textbf{84.82} & \textbf{92.57} & \textbf{57.45} & \textbf{92.43} \\
 & E$\,\to\,$M & 92.67 & 41.43 & 27.76 & 70.35 & \textbf{21.43} & 39.98 & 89.44 & 79.79 & 48.4 & 83.19 & 88.88 & 54.72 & 90.05 \\ \cline{2-15}
 & E$\,\to\,$E & 89.75 & 36.17 & 35.10 & 70.38 & 27.50 & 42.89 & 88.79 & 73.67 & 50.14 & 73.55 & 84.82 & 55.80 & 84.83 \\\hline \hline
\multirow{7}{*}{ContextBERT} & D$\,\to\,$D & 80.16 & 0 & \textbf{75.69} & 0 & 5 & 58.58 & \textbf{71.69} & \textbf{73.91} & 35.16 & 72.85 & 77.19 & 41.59 & 76.81 \\
 & M$\,\to\,$D & 72.11 & \textbf{11.67} & 7.73 & \textbf{43.71} & 11.87 & \textbf{60.07} & 66.29 & 21.29 & 33.56 & 14.49 & 57.80 & 39.06 & 45.67 \\
 & E$\,\to\,$D & \textbf{81.58} & 5.00 & 75.46 & 40.00 & \textbf{52.81} & 57.31 & 69.23 & 73.71 & \textbf{49.97} & \textbf{73.49} & \textbf{77.89} & \textbf{54.48} & \textbf{77.87} \\ \cline{2-15}
 & D$\,\to\,$M & 89.37 & 0 & 11.18 & 0 & 0 & 13.86 & 77.07 & 59.43 & 17.02 & 67.81 & 80.44 & 27.35 & 83.40 \\
 & M$\,\to\,$M & \textbf{95.09} & \textbf{35.71} & \textbf{36.35} & \textbf{70.34} & \textbf{19.44} & 34.05 & \textbf{90.01} & \textbf{84.36} & \textbf{47.65} & \textbf{83.87} & \textbf{92.14} & \textbf{54.43} & \textbf{91.98} \\
 & E$\,\to\,$M & 93.45 & 33.70 & 30.39 & 62.42 & 17.27 & \textbf{37.06} & 89.75 & 80.44 & 45.10 & 83.14 & 89.65 & 52.00 & 90.60 \\ \cline{2-15}
 & E$\,\to\,$E & 92.10 & 30.08 & 61.69 & 62.36 & 41.73 & 40.83 & 89.14 & 78.99 & 54.30 & 79.67 & 87.93 & 59.70 & 88.33 \\ \hline \hline
\multirow{7}{*}{\begin{tabular}[c]{@{}l@{}}DialogueRNN\\ (GloVe)\end{tabular}} & D$\,\to\,$D & 40.13 & 0 & \textbf{64.01} & 0 & 0 & 52.05 & 65.59 & \textbf{62.56} & 30.28 & \textbf{62.03} & \textbf{54.88} & 31.68 & \textbf{50.18} \\
 & M$\,\to\,$D & \textbf{67.00} & 0 & 22.91 & \textbf{100} & 0 & 54.45 & 55.96 & 31.03 & 38.89 & 26.23 & 54.07 & \textbf{42.90} & 48.30 \\
 & E$\,\to\,$D & 23.83 & 0 & 61.75 & 60 & 0 & \textbf{63.72} & \textbf{73.57} & 62.06 & \textbf{43.17} & 61.23 & 50.24 & 40.41 & 40.99 \\ \cline{2-15}
 & D$\,\to\,$M & 59.78 & 0 & 5.34 & 0 & 0 & 13.80 & 85.57 & 50.51 & 17.45 & 74.87 & 55.46 & 23.50 & 63.96 \\
 & M$\,\to\,$M & 87.25 & \textbf{21.57} & \textbf{21.53} & 52.16 & 0 & 26.21 & 85.51 & 72.78 & \textbf{34.50} & 78.12 & 82.21 & \textbf{42.03} & 84.72 \\
 & E$\,\to\,$M & \textbf{88.24} & 13.59 & 18.67 & \textbf{57.56} & 0 & \textbf{27.92} & \textbf{86.73} & \textbf{74.04} & 34.08 & \textbf{79.22} & \textbf{83.41} & 41.81 & \textbf{85.74} \\ \cline{2-15}
 & E$\,\to\,$E & 83.46 & 12.71 & 51.38 & 57.67 & 0 & 32.75 & 86.35 & 70.93 & 40.14 & 74.56 & 78.56 & 46.33 & 80.76 \\ \hline \hline
\multirow{7}{*}{\begin{tabular}[c]{@{}l@{}}DialogueRNN\\ (BERT)\end{tabular}} & D$\,\to\,$D & 65.24 & 0 & 58.24 & 0 & 27.69 & 54.51 & \textbf{68.35} & 58.45 & 34.80 & 57.97 & \textbf{61.95} & 39.15 & \textbf{61.90} \\
 & M$\,\to\,$D & \textbf{66.63} & 0 & 4.24 & 43.52 & 2.86 & 41.48 & 53.87 & 17.33 & 24.33 & 9.64 & 50.73 & 30.37 & 40.48 \\
 & E$\,\to\,$D & 49.81 & 0 & \textbf{61.01} & \textbf{91.67} & \textbf{28.14} & \textbf{60.92} & 66.70 & \textbf{60.95} & \textbf{51.41} & \textbf{60.56} & 56.58 & \textbf{51.18} & 54.74 \\ \cline{2-15}
 & D$\,\to\,$M & 85.48 & 0 & 8.17 & 0 & 6.71 & 20.48 & 87.46 & 65.74 & 20.47 & 76.91 & 78.71 & 29.76 & 83.11 \\
 & M$\,\to\,$M & \textbf{92.11} & 34.83 & \textbf{34.49} & 58.59 & 0 & 26.32 & 87.48 & \textbf{79.18} & 40.28 & 80.84 & \textbf{88.13} & 47.69 & \textbf{88.99} \\
 & E$\,\to\,$M & 90.54 & \textbf{47.12} & 18.08 & \textbf{71.22} & \textbf{15.48} & \textbf{33.92} & \textbf{88.28} & 76.26 & \textbf{45.68} & \textbf{81.52} & 86.10 & \textbf{52.09} & 88.04 \\ \cline{2-15}
 & E$\,\to\,$E & 86.85 & 41.32 & 47.51 & 71.48 & 25.56 & 39.42 & 87.58 & 72.48 & 52.15 & 75.50 & 81.78 & 57.10 & 83.41 \\ \hline \hline
\multirow{7}{*}{COSMIC} & D$\,\to\,$D & 69.34 & 0 & 59.68 & 0 & 0 & 64.30 & \textbf{72.25} & 60.31 & 32.71 & 59.25 & \textbf{65.07} & 37.94 & \textbf{64.71} \\
 & M$\,\to\,$D & \textbf{71.56} & 33.33 & 2.67 & 100 & 15.38 & 67.04 & 70.80 & 19.98 & 48.21 & 11.03 & 57.16 & 51.54 & 43.79 \\
 & E$\,\to\,$D & 66.59 & 0 & \textbf{61.47} & \textbf{100} & \textbf{43.71} & \textbf{69.74} & 68.19 & \textbf{62.09} & \textbf{57.18} & \textbf{61.67} & 64.47 & \textbf{58.53} & 64.33 \\ \cline{2-15}
 & D$\,\to\,$M & 86.68 & 0 & 8.78 & 0 & 0 & 20.91 & 88.90 & 67.85 & 19.77 & 78.19 & 80.32 & 29.32 & 84.33 \\
 & M$\,\to\,$M & \textbf{94.86} & 50 & \textbf{40.97} & 67.12 & 0 & 41.77 & \textbf{89.93} & \textbf{84.22} & \textbf{48.30} & \textbf{84.27} & \textbf{91.81} & \textbf{54.95} & \textbf{91.93} \\
 & E$\,\to\,$M & 92.61 & \textbf{58.18} & 24.68 & \textbf{70.52} & \textbf{0} & \textbf{37.92} & 89.10 & 79.84 & 46.73 & 82.74 & 88.81 & 53.29 & 89.88 \\ \cline{2-15}
 & E$\,\to\,$E & 89.80 & 51.98 & 50.69 & 70.93 & 31.62 & 44.42 & 88.42 & 75.89 & 56.34 & 77.09 & 85.26 & 61.12 & 85.94 \\
 \bottomrule[1pt]
\end{tabular}
    \caption{Performance of baseline models on emotion classification including cross-dataset experiments. For cross-dataset experiments, the ``X$\,\to\,$Y''s in the 'Set-up' column represents the training and evaluation set-up, where X is the training set and Y is the test set. E stands for EmoWOZ, M stands for MultiWOZ, and D stands for DialMAGE. M$\,\to\,$D, for example, means to train on MultiWOZ and test on DialMAGE. Extreme values for ``Apologetic'' and ``Abusive'' in DialMAGE (``*$\,\to\,$D''s) are caused by their rarity in the test set (1 and 5 occurrences respectively).}
    \label{tab:f1-cross-emotion}
\end{table}

\subsection{Confusion Matrix of ContextBERT}
\begin{figure}[H]
\centering
\includegraphics[width=0.4\textwidth]{{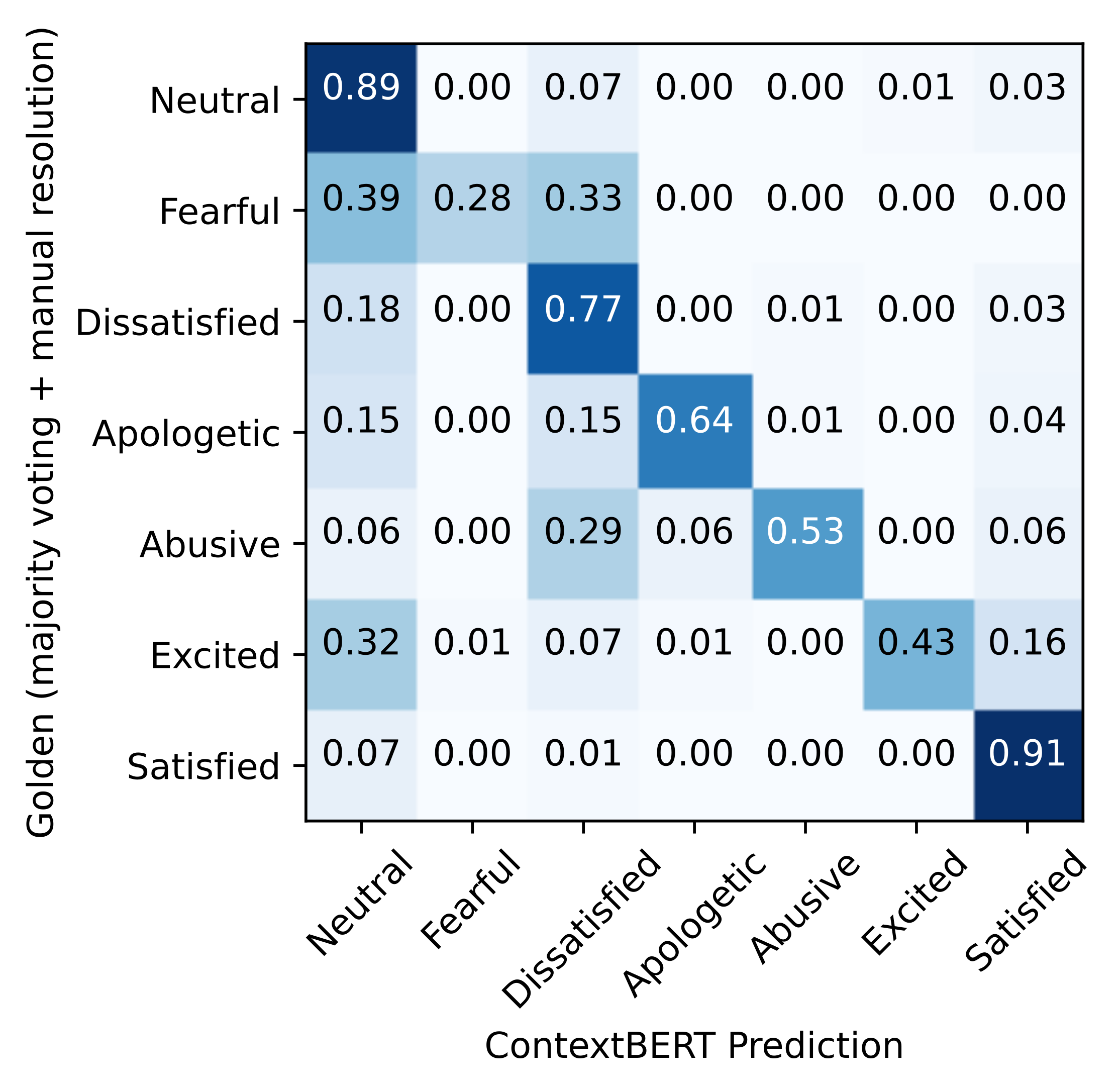}}
\caption{Confusion matrix between Golden Labels and (the best) ContextBERT Prediction.}
\label{fig:contextbert-cm}
\end{figure}

\newpage
\subsection{Sentiment Classification (3 classes)}
\label{sec:sentiment-classification-results}
\begin{table}[H]
\centering
\small
\setlength\tabcolsep{2pt}
\begin{tabular}{ccc|ccc|cc|cc|cc}
\toprule[1pt]
\multirow{3}{*}{\textbf{Model}} & \multirow{3}{*}{\textbf{Feature}} & \multirow{3}{*}{\textbf{Ctx.}} & \multicolumn{3}{c|}{\multirow{2}{*}{\textbf{\begin{tabular}[c]{@{}c@{}}F1 of Each Sentiment\\ in EmoWOZ\end{tabular}}}} & \multicolumn{6}{c}{\textbf{Average F1 w/o Neutral}} \\
 &  &  & \multicolumn{3}{c|}{} & \multicolumn{2}{c|}{EmoWOZ} & \multicolumn{2}{c|}{MultiWOZ} & \multicolumn{2}{c}{DialMAGE} \\
 &  &  & Neutral & Negative & Positive & Macro & Weighted & Macro & Weighted & Macro & Weighted \\ \hline
BERT & BERT & No & 89.9 & 42.0 & 88.3 & 65.1 & 75.6 & 67.0 & 84.6 & 54.8 & 44.7 \\
ContextBERT & BERT & Yes & \textbf{92.7} & \textbf{68.0} & 87.5 & \textbf{77.8} & \textbf{82.2} & \textbf{70.2} & \textbf{84.7} & \textbf{71.2} & \textbf{75.3} \\
DialogueRNN & GloVe & Yes & 86.8 & 61.0 & 83.7 & 72.4 & 77.5 & 66.5 & 80.6 & 69.0 & 66.7 \\
DialogueRNN & BERT & Yes & 83.3 & 47.2 & 87.4 & 67.3 & 76.4 & 57.0 & 81.8 & 67.2 & 66.1 \\
COSMIC & BERT+COMET & Yes & 89.0 & 48.2 & \textbf{88.5} & 68.3 & 77.5 & 63.0 & 84.0 & 63.7 & 59.1 \\
\bottomrule[1pt]
\end{tabular}
\caption{Summarised performance of baseline models on sentiment classification. ``Ctx.'' stands for ``context''. \label{tab:f1-baseline-sentiment}}
\end{table}

\begin{table}[H]
\centering
\small
\setlength\tabcolsep{4pt}

\begin{tabular}{l|c|ccc|ccc|ccc}
\toprule[1pt]
\multirow{2}{*}{\textbf{Model}} & \multirow{2}{*}{\textbf{Set-up}} & \multicolumn{3}{l|}{\textbf{F1 for each Sentiment Label}} & \multicolumn{3}{l|}{\textbf{Average F1 w/o Neutral}} & \multicolumn{3}{l}{\textbf{Average F1 w Neutral}} \\ 
 &  & Neutral & Negative & Positive & Micro & Macro & Weighted & Micro & Macro & Weighted \\\hline
\multirow{7}{*}{BERT} & D$\,\to\,$D & 57.61 & \textbf{61.32} & \textbf{72.48} & \textbf{62.45} & \textbf{66.90} & \textbf{62.61} & 60.54 & \textbf{63.80} & \textbf{59.91} \\
 & M$\,\to\,$D & \textbf{71.70} & 3.80 & 68.47 & 17.78 & 36.13 & 11.27 & 57.39 & 47.99 & 43.93 \\
 & E$\,\to\,$D & 70.01 & 41.73 & 67.84 & 46.16 & 54.78 & 44.74 & \textbf{61.68} & 59.86 & 58.39 \\ \cline{2-11}
 & D$\,\to\,$M & 71.97 & 7.77 & 84.71 & 56.23 & 46.24 & 76.94 & 65.82 & 54.82 & 73.35 \\
 & M$\,\to\,$M & \textbf{95.43} & \textbf{56.68} & \textbf{90.05} & \textbf{87.00} & \textbf{73.37} & \textbf{86.69} & \textbf{93.11} & \textbf{80.72} & \textbf{92.99} \\
 & E$\,\to\,$M & 92.90 & 44.99 & 89.08 & 81.99 & 67.04 & 84.63 & 89.66 & 75.66 & 90.60 \\ \cline{2-11}
 & E$\,\to\,$E & 89.93 & 41.96 & 88.27 & 75.74 & 65.11 & 75.60 & 85.57 & 73.39 & 85.56 \\\hline\hline
\multirow{7}{*}{ContextBERT} & D$\,\to\,$D & 80.67 & \textbf{76.74} & \textbf{70.81} & \textbf{76.08} & \textbf{73.77} & \textbf{76.05} & 78.53 & \textbf{76.07} & 78.55 \\
 & M$\,\to\,$D & 71.69 & 6.65 & 64.22 & 20.02 & 35.43 & 13.30 & 57.27 & 47.52 & 44.85 \\
 & E$\,\to\,$D & \textbf{82.91} & 76.60 & 65.71 & 75.25 & 71.15 & 75.34 & \textbf{79.48} & 75.07 & \textbf{79.43} \\\cline{2-11}
 & D$\,\to\,$M & 89.04 & 17.89 & 69.53 & 54.85 & 43.71 & 64.32 & 79.01 & 58.82 & 82.16 \\
 & M$\,\to\,$M & \textbf{95.38} & \textbf{55.09} & \textbf{89.89} & \textbf{86.81} & \textbf{72.49} & \textbf{86.38} & \textbf{93.03} & \textbf{80.12} & \textbf{92.88} \\
 & E$\,\to\,$M & 93.98 & 51.94 & 88.38 & 83.75 & 70.16 & 84.70 & 91.05 & 78.10 & 91.40 \\ \cline{2-11}
 & E$\,\to\,$E & 92.69 & 67.98 & 87.55 & 81.95 & 77.76 & 82.19 & 89.36 & 82.74 & 89.49 \\\hline\hline
\multirow{7}{*}{\begin{tabular}[c]{@{}l@{}}DialogueRNN\\ (GloVe)\end{tabular}} 
 & D$\,\to\,$D & 58.50 & \textbf{69.49} & 67.55 & \textbf{69.21} & 68.52 & 69.26 & \textbf{64.98} & \textbf{65.18} & \textbf{63.45} \\
 & M$\,\to\,$D & \textbf{68.70} & 20.44 & 52.21 & 29.70 & 36.32 & 24.11 & 55.21 & 47.12 & 48.21 \\
 & E$\,\to\,$D & 42.57 & 66.00 & \textbf{71.96} & 66.56 & \textbf{68.98} & \textbf{66.68} & 58.16 & 60.17 & 53.65 \\ \cline{2-11}
 & D$\,\to\,$M & 77.27 & 10.88 & \textbf{84.93} & 62.22 & 47.90 & 77.46 & 71.38 & 57.69 & 77.32 \\
 & M$\,\to\,$M & \textbf{91.06} & \textbf{50.36} & 84.13 & \textbf{80.22} & \textbf{67.25} & \textbf{80.72} & \textbf{87.57} & \textbf{75.19} & \textbf{88.18} \\
 & E$\,\to\,$M & 90.71 & 48.85 & 84.21 & 79.77 & 66.53 & 80.64 & 87.14 & 74.59 & 87.91 \\ \cline{2-11}
 & E$\,\to\,$E & 86.75 & 61.03 & 83.74 & 76.42 & 72.39 & 77.53 & 82.91 & 77.18 & 83.94 \\\hline\hline
\multirow{7}{*}{\begin{tabular}[c]{@{}l@{}}DialogueRNN\\(BERT)\end{tabular}} & D$\,\to\,$D & 47.53 & 65.27 & \textbf{69.53} & 65.74 & \textbf{67.40} & 65.76 & 59.05 & 60.78 & 55.91 \\
 & M$\,\to\,$D & \textbf{71.10} & 7.04 & 64.41 & 20.25 & 35.72 & 13.67 & 56.77 & 47.52 & 44.70 \\
 & E$\,\to\,$D & 50.18 & \textbf{65.81} & 68.63 & \textbf{66.10} & 67.22 & \textbf{66.14} & \textbf{60.04} & \textbf{61.54} & \textbf{57.52} \\\cline{2-11}
 & D$\,\to\,$M & 59.37 & 8.08 & 86.36 & 51.21 & 47.22 & 78.46 & 55.78 & 51.27 & 64.68 \\
 & M$\,\to\,$M & \textbf{93.75} & \textbf{52.28} & \textbf{88.13} & \textbf{84.27} & \textbf{70.20} & \textbf{84.52} & \textbf{90.94} & \textbf{78.05} & \textbf{91.18} \\
 & E$\,\to\,$M & 86.61 & 25.98 & 88.11 & 73.03 & 57.04 & 81.84 & 81.92 & 66.90 & 85.29 \\ \cline{2-11}
 & E$\,\to\,$E & 83.30 & 47.16 & 87.36 & 71.40 & 67.26 & 76.36 & 78.72 & 72.61 & 81.19 \\\hline\hline
\multirow{7}{*}{COSMIC} & D$\,\to\,$D & 63.51 & \textbf{68.32} & \textbf{70.52} & \textbf{68.58} & \textbf{69.42} & \textbf{68.57} & \textbf{66.45} & \textbf{67.45} & \textbf{65.83} \\
 & M$\,\to\,$D & \textbf{71.69} & 5.22 & 67.57 & 18.79 & 36.39 & 12.42 & 57.32 & 48.16 & 44.45 \\
 & E$\,\to\,$D & 69.28 & 57.68 & 69.66 & 59.30 & 63.67 & 59.06 & 64.80 & 65.54 & 64.58 \\\cline{2-11}
 & D$\,\to\,$M & 80.31 & 12.43 & 86.38 & 62.62 & 49.41 & 78.92 & 73.66 & 59.71 & 79.92 \\
 & M$\,\to\,$M & \textbf{95.01} & \textbf{57.85} & \textbf{89.48} & \textbf{86.41} & \textbf{73.67} & \textbf{86.29} & \textbf{92.58} & \textbf{80.78} & \textbf{92.58} \\
 & E$\,\to\,$M & 91.63 & 36.68 & 89.32 & 79.78 & 63.00 & 84.01 & 87.99 & 72.54 & 89.51 \\ \cline{2-11}
 & E$\,\to\,$E & 88.97 & 48.15 & 88.55 & 75.66 & 68.35 & 77.5 & 84.6 & 75.22 & 85.47 \\
\bottomrule[1pt]
\end{tabular}
\caption{Detailed results of baseline models on sentiment classification including cross-dataset experiments. For cross-dataset experiments, the ``X$\,\to\,$X''s in the 'Set-up' column represents the training and evaluation set-up. E stands for EmoWOZ, M stands for MultiWOZ, and D stands for DialMAGE. M$\,\to\,$D, for example, means to train on MultiWOZ and test on DialMAGE.}
    \label{tab:f1-cross-sentiment}

\end{table}

\newpage
\subsection{Change in precision and recall on MultiWOZ after Complementing MultiWOZ with DialMAGE in Training}
\label{sec:precision-recall}

\begin{table}[H]
\centering
\footnotesize
\setlength\tabcolsep{3pt}
\begin{tabular}{l|l|ccccccc}
\toprule[1pt]
 &  & Neutral & Fearful & Dissatisfied & Apologetic & Abusive & Excited & Satisfied \\ \hline
\multirow{3}{*}{ContextBERT} & Recall & \textbf{\color{red} 95.3$\,\to\,$91.5} & \textbf{\color{red} 34.7$\,\to\,$28.0} & \textbf{\color[HTML]{009b55} 31.4$\,\to\,$60.4} & 69.7$\,\to\,$61.9 & 16.0$\,\to\,$24.0 & 33.5$\,\to\,$34.1 & 90.4$\,\to\,$90.0 \\
 & Precision & \textbf{\color[HTML]{009b55}94.9$\,\to\,$95.5} & \textbf{\color[HTML]{009b55} 37.3$\,\to\,$44.8} & \textbf{\color{red} 43.7$\,\to\,$20.9} & 71.4$\,\to\,$63.5 & 25.0$\,\to\,$13.7 & \textbf{\color[HTML]{009b55} 35.5$\,\to\,$42.6} & 89.5$\,\to\,$89.6 \\
 & F1 & \textbf{\color{red} 95.1$\,\to\,$93.5} & 35.7$\,\to\,$33.7 & 36.4$\,\to\,$30.4 & \textbf{\color{red} 70.3$\,\to\,$62.4} & 19.4$\,\to\,$17.3 & 34.0$\,\to\,$37.1 & 90.0$\,\to\,$89.7 \\
 \bottomrule[1pt]
\end{tabular}
\caption{Precision, recall and F1 score of ContextBERT for all emotions when trained on MultiWOZ and EmoWOZ respectively, and tested on MultiWOZ. $A\,\to\,B$ represents how the value change after complementing MultiWOZ with DialMAGE in training. $A$ is the value when trained on MultiWOZ and $B$ is the value when trained on EmoWOZ. Values with statistical significance ($p < 0.1$) are bolded and colored where red indicates a drop and green indicates an improvement. For recognising user emotions in task-oriented dialogues, a high precision is more desirable for \textit{neutral}, \textit{fearful}, \textit{apologetic}, \textit{abusive}, \textit{excited}, and \textit{satisfied} where as a high recall is more desirable for \textit{dissatisfied}. \label{tab:precision-recall-full}}

\end{table}

\subsection{Emotion Distribution in Model Predictions}
\label{sec:emo-dist-model-pred}

\begin{table}[H]
\centering
\small
\setlength\tabcolsep{5pt}
\begin{tabular}{l|l|ccccccc}
\toprule[1pt]
\multicolumn{1}{c}{Test Set} & \multicolumn{1}{c}{Model} & Neutral & Fearful & Dissatisfied & Apologetic & Abusive & Excited & Satisfied \\ \hline
\multicolumn{2}{l|}{MultiWOZ Label} & 72.31 & 0.2 & 1.47 & 0.98 & 0.07 & 1.0 & 23.97 \\
\multicolumn{2}{l|}{DialMAGE (\#token\textgreater{}11.8) Label} & 61.96 & 0.61 & 28.83 & 0.61 & 0.0 & 6.75 & 1.23 \\ \hline
\multirow{5}{*}{\begin{tabular}[c]{@{}l@{}}DialMAGE \\ (\#token \textgreater{} 11.8) \\ Prediction\end{tabular}} & BERT & 89.57 & 0.0 & 2.45 & 0.0 & 0.0 & 6.75 & 1.23 \\
 & ContextBERT & 58.13 & 0.0 & 35.47 & 0.0 & 0.0 & 4.93 & 1.48 \\
 & DialogueRNN-GloVe & 6.52 & 0.0 & 74.97 & 1.28 & 0.0 & 6.05 & 11.18 \\
 & DialogueRNN-BERT & 55.65 & 0.12 & 23.05 & 8.61 & 0.0 & 5.01 & 7.57 \\
 & COSMIC & 61.93 & 0.23 & 18.63 & 7.57 & 0.0 & 3.84 & 7.8 \\ \hline
\multicolumn{2}{l|}{DialMAGE Label} & 54.12 & 0.24 & 39.3 & 0.08 & 0.95 & 1.35 & 3.96 \\
\multicolumn{2}{l|}{MultiWOZ (\#token\textless{}5.8) Label} & 60.76 & 0.0 & 1.21 & 0.0 & 0.0 & 0.3 & 37.73 \\ \hline
\multirow{5}{*}{\begin{tabular}[c]{@{}l@{}}MultiWOZ \\ (\#token \textless{} 5.8) \\ Prediction\end{tabular}} & BERT & 60.91 & 0.15 & 0.45 & 0.3 & 0.0 & 0.45 & 37.73 \\
 & ContextBERT & 57.43 & 0.0 & 2.97 & 0.0 & 0.0 & 0.66 & 38.94 \\
 & DialogueRNN-GloVe & 46.54 & 0.0 & 2.1 & 0.7 & 0.0 & 0.88 & 49.78 \\
 & DialogueRNN-BERT & 46.28 & 0.0 & 8.94 & 0.26 & 0.09 & 1.67 & 42.77 \\
 & COSMIC & 49.43 & 0.09 & 5.0 & 0.18 & 0.09 & 1.58 & 43.65 \\
\bottomrule[1pt]
\end{tabular}
\caption{Emotion distribution in model predictions (trained on EmoWOZ).  \label{tab:model-prediction-distribution}}
\end{table}

\end{document}